\documentclass{article}

\usepackage{microtype}
\usepackage{graphicx}
\usepackage{subcaption}
\usepackage{booktabs}
\usepackage{algorithm}
\usepackage{enumitem}
\usepackage{algpseudocode}
\usepackage{svg}
\usepackage{xcolor}

\definecolor{firebrick}{RGB}{178,34,34}

\usepackage{hyperref}

\usepackage[preprint]{icml2026}

\usepackage{amsmath}
\usepackage{amssymb}
\usepackage{mathtools}
\usepackage{amsthm}
\usepackage{multirow}
\usepackage[capitalize,noabbrev]{cleveref}

\theoremstyle{plain}

\theoremstyle{definition}

\theoremstyle{remark}

\usepackage[textsize=tiny]{todonotes}

\AtBeginDocument{
  \hypersetup{
    colorlinks=true,
    linkcolor=firebrick,
    citecolor=firebrick,
    urlcolor=firebrick,
  }
}

\icmltitlerunning{Spectral Collapse in Diffusion Inversion}

\begin{document}
\twocolumn[
  \icmltitle{Spectral Collapse in Diffusion Inversion}



  \icmlsetsymbol{equal}{*}

  \begin{icmlauthorlist}
    \icmlauthor{Nicolas Bourriez}{equal,ens}
    \icmlauthor{Alexandre Vérine}{ens}
    \icmlauthor{Auguste Genovesio}{ens}
  \end{icmlauthorlist}

  \icmlaffiliation{ens}{Ecole Normale Supérieure PSL, Paris, France}

  \icmlcorrespondingauthor{Auguste Genovesio}{auguste.genovesio@ens.psl.eu}

  \icmlkeywords{Diffusion, Inversion, Score-Matching, Machine Learning, ICML}

  \vskip 0.3in
]



\printAffiliationsAndNotice{}  

\begin{abstract}
  Conditional diffusion inversion provides a powerful framework for unpaired image-to-image translation. However, we demonstrate through an extensive analysis that standard deterministic inversion (e.g. DDIM) fails when the source domain is spectrally sparse compared to the target domain (e.g., super-resolution, sketch-to-image). In these contexts, the recovered latent from the input does not follow the expected isotropic Gaussian distribution. Instead it exhibits a signal with lower frequencies, locking target sampling to oversmoothed and texture-poor generations. We term this phenomenon \textit{spectral collapse}.
  We observe that stochastic alternatives attempting to restore the noise variance tend to break the semantic link to the input, leading to structural drift. To resolve this structure-texture trade-off, we propose Orthogonal Variance Guidance (OVG), an inference-time method that corrects the ODE dynamics to enforce the theoretical Gaussian noise magnitude within the null-space of the structural gradient. Extensive experiments on microscopy super-resolution (BBBC021) and sketch-to-image (Edges2Shoes) demonstrate that OVG effectively restores photorealistic textures while preserving structural fidelity.
\end{abstract}
\section{Introduction}
\label{sec:intro}

\begin{figure}[ht!]
    \centering
    \includegraphics[width=1.0\columnwidth]{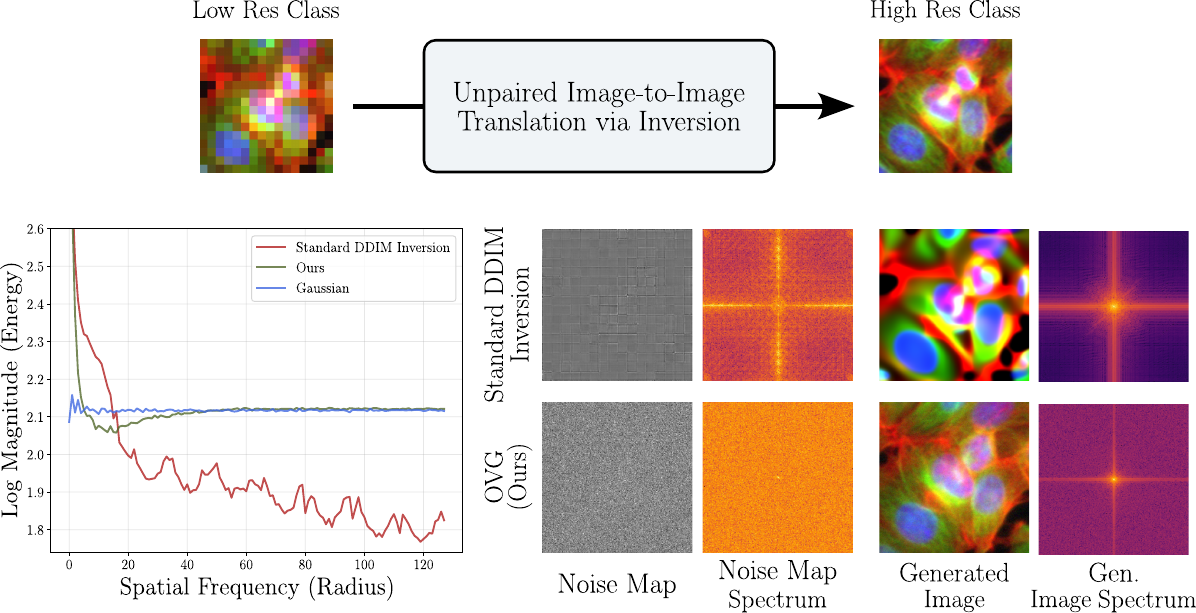}
    \caption{\textbf{Spectral collapse.} Unpaired image-to-image translation via diffusion inversion fails when the source domain is spectrally sparse (e.g., Low Res $\to$ High Res). Inverted noise maps display significantly reduced high-frequency energy compared to the theoretical Gaussian prior and retain low-frequency structural imprints. Such \textit{spectrally collapsed} noise maps lead to oversmoothed generations in the target domain that lack texture. Our proposed method restores the correct Gaussian spectral profile, enabling the generation of realistic high-frequency textures, while preserving structural fidelity.}
    \label{fig:spectral_collapse_psd}
\end{figure}

Denoising Diffusion Probabilistic Models (DDPMs) \cite{ho2020denoisingdiffusionprobabilisticmodels, kawar2022denoisingdiffusionrestorationmodels, song2021scorebasedgenerativemodelingstochastic} have established themselves as the state-of-the-art for high-fidelity image synthesis, progressively displacing GANs due to their stable training and mode-covering capabilities. Beyond unconditional generation, their formulation as a reversible probability flow has made them a powerful tool for inverse problems, enabling zero-shot editing, inpainting, and restoration via "Plug-and-Play" guidance \cite{chung2022diffusionposteriorsamplinggeneral,meng2021sdeditguidedimagesynthesis}.
Central to these editing workflows is the concept of \textit{inversion}: mapping a real image $\mathbf{x}_0$ back to its latent noise representation $\mathbf{z}_T$ such that the generative process can reconstruct it.
Considerable research effort has been directed towards perfecting this reconstruction. Methods like EDICT \cite{wallace2022edictexactdiffusioninversion} or BDIA \cite{zhang2023exactdiffusioninversionbidirectional} have achieved near-exact pixel-level recovery by correcting the discretization errors of the deterministic PF-ODE.
However, these approaches focus on correcting \textit{reconstruction} errors, not \textit{generating} new details.

\begin{figure*}[t]
    \centering
    \includegraphics[width=1.0\textwidth]{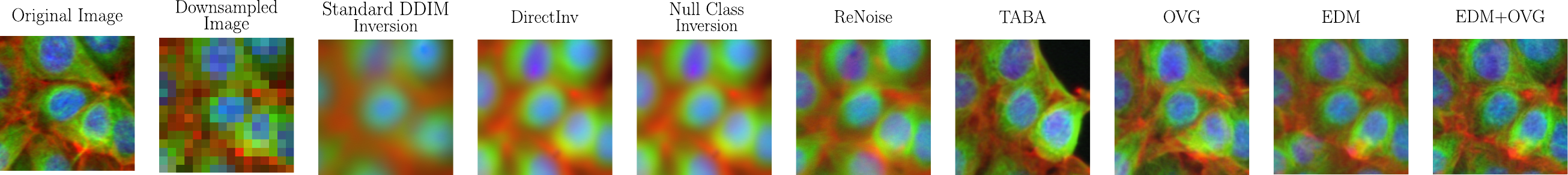}
    \caption{Qualitative comparison on BBBC021 ($\times 16$ super-resolution) using models trained in latent space.
        Deterministic baselines (DDIM, DirectInv, Null-Class) suffer from \textit{spectral collapse}, yielding oversmoothed outputs.
        Stochastic methods (ReNoise, TABA) and standard EDM exhibit artifacts or structural drift.
        Our EDM+OVG approach uniquely restores realistic high-frequency texture while maintaining strict structural fidelity to the input.}
    \label{fig:comparison_methods_grid}
\end{figure*}

In this work, we address an issue that cannot be solved by mitigating reconstruction error: it occurs in tasks such as unpaired image-to-image translation from a spectrally \textit{sparse} to a spectrally \textit{dense} domain. It is striking in tasks such as Super-Resolution (e.g., microscopy) or sketch-to-image, where the input image contains strictly less information than the desired output image. In many real-world cases, we have no access to paired samples and no plausible downsampling function that allows one to browse the latent space (such as in \cite{chung2022diffusionposteriorsamplinggeneral}). In these cases, the objective  not only consists of reconstructing the input's low-frequency "imprint", but also to hallucinate plausible high-frequency textures that do not exist in the source $\mathbf{x}_0$.
We observe that in these sparse to dense image-to-image translation tasks, standard deterministic inversion suffers from a phenomenon we term \textit{spectral collapse}: the recovered latent noise map $\mathbf{z}_T$ is \textit{not} an isotropic Gaussian but a low-frequency structural residual of the input. Because this "collapsed" latent lacks the stochastic variance required to generate new textures, conditional sampling results in \textit{poor high-frequencies} outputs.

Existing attempts to fix this rely on stochastic noise injection, such as in "There and Back Again" (TABA) \cite{staniszewski2025againrelationnoiseimage}. While successful at restoring texture, these methods break the intrinsic link between the noise map and the input structure, leading to "structural hallucinations" where the generated content drifts semantically away from the source.

We propose to solve this by rigorously analyzing the spectral properties of the inversion trajectory, and proposing an adapted plug-and-play method to guide inversion at inference. Code can be found at \href{https://github.com/nicoboou/ovg}{github.com/nicoboou/ovg}. Our contributions are as follows:
\begin{itemize}[topsep=0pt, leftmargin=*]
    \item We perform a systematic empirical analysis of diffusion inversion across diverse \textit{spectrally sparse} domains (Super Resolution, sketch-to-image), identifying \textbf{spectral collapse} as the primary bottleneck for translation quality.
    \item We demonstrate that the $\mathbf{x}_0$-prediction training objective via EDM framework \cite{karras2022elucidatingdesignspacediffusionbased} is inherently more robust to this collapse than the classical Denoising Diffusion Implicit Model (DDIM) $\boldsymbol{\epsilon}$-prediction regime.~\cite{song2020denoisingdiffusionimplicitmodels}.
    \item We introduce \textbf{Orthogonal Variance Guidance (OVG)}, a guidance method that injects high-frequency variance strictly within the null-space of the structural condition from the source. By ensuring orthogonality at inference, OVG restores Gaussianity required for realistic texture without corrupting the semantic layout of the input.
\end{itemize}
\section{Background}
\label{sec:background}

\subsection{Probability Flow ODE}
\label{subsec:pf_ode}
We consider a \emph{conditional} data distribution $p_0(\mathbf{x}\mid y)$, where $y\in\mathcal{Y}$ denotes a domain-, task-, or label-specific variable (e.g., class label, text prompt embedding, or measurement context). Diffusion models define a stochastic forward process that gradually corrupts clean samples $\mathbf{x}_0\sim p_0(\cdot\mid y)$ into Gaussian noise at terminal time $t=T$. In continuous-time formulation, this corruption is a SDE whose marginal densities $\{p_t(\mathbf{x}\mid y)\}_{t=0}^T$ are shared by a deterministic \emph{Probability Flow ODE} (PF-ODE) \cite{song2021scorebasedgenerativemodelingstochastic}.

While diffusion sampling can be performed by integrating either the stochastic reverse-time SDE or the deterministic PF-ODE, image-to-image translation necessitates a bijective mapping between observations and latents. Consequently, we adopt the deterministic PF-ODE formulation for our inversion and sampling pipeline. In its standard form, this ODE is driven by the conditional score $\nabla_{\mathbf{x}}\log p_t(\mathbf{x}\mid y)$:
\begin{equation}
    d\mathbf{x} = \left[ f(t)\mathbf{x} - \frac{1}{2} g(t)^2\, \nabla_{\mathbf{x}} \log p_t(\mathbf{x}\mid y) \right] dt,
    \label{eq:standard_pf_ode}
\end{equation}
where $f(t)$ and $g(t)$ denote the drift and diffusion coefficients of the corresponding forward SDE.

Modern diffusion models parameterize the score function using a neural network $F_\theta(\mathbf{x}_t, y, t)$. Depending on the training objective, this network may predict the noise $\boldsymbol{\epsilon}$, the clean signal $\mathbf{x}_0$, or a velocity variable $v$, all being linearly related to the score. Regardless of this choice, a wide class of deterministic samplers can be written as an \emph{affine} update along the PF-ODE trajectory (from $t=T$ to $t=0$):
\begin{equation}
    \mathbf{x}_{t-1} = a(t)\,\mathbf{x}_t + b(t)\,F_\theta(\mathbf{x}_t, y, t),
    \label{eq:general_ode_backward}
\end{equation}
where the scalar schedules $a(t)$ and $b(t)$ depend on the chosen parameterization and discretization scheme.

\subsection{Inversion}\label{subsec:inversion}
In inversion, we integrate the deterministic flow forward in time to map an input $\mathbf{x}_0$ to a latent $\mathbf{x}_T$ that should follow the model prior. Using the affine form in Eq.~\ref{eq:general_ode_backward}, a natural forward-time inversion step (from $t$ to $t+1$) is obtained by algebraic inversion of the affine map:
\begin{equation}
    \mathbf{x}_{t+1} = \frac{\mathbf{x}_t - b(t)\,F_\theta(\mathbf{x}_t, \bar{y}, t)}{a(t)},
    \label{eq:general_ode_forward}
\end{equation}
where $\bar{y}$ denotes the conditioning associated with the input (source) domain.

A common editing/transfer procedure then consists of: (i) \emph{invert} $\mathbf{x}_0$ under $\bar{y}$ to obtain a latent $\mathbf{x}_T$, and (ii) \emph{generate} from the same $\mathbf{x}_T$ under a target conditioning $y$ by integrating Eq.~\ref{eq:general_ode_backward} back to $t=0$. This yields a deterministic label-to-label mapping $\bar{y}\to y$ mediated by the shared latent.

\paragraph{The linear assumption}
The forward inversion update in Eq.~\ref{eq:general_ode_forward} is exact only in the infinitesimal-step limit. In practice, discretization implicitly assumes that the vector field does not change appreciably across a step. Following \cite{wallace2022edictexactdiffusioninversion}, this is commonly summarized as a \emph{linear assumption}:
\begin{equation}
    F_\theta(\mathbf{x}_{t}, y, t) \approx F_\theta(\mathbf{x}_{t-1}, y, t)
    \label{eq:linear_assumption}
\end{equation}
where the time index is fixed to isolate the spatial non-linearity.
When Eq.~\ref{eq:linear_assumption} is violated, forward-time inversion no longer traces the true reverse trajectory of the deterministic sampler. The resulting drift mismatch accumulates over steps, so the terminal latent $\mathbf{x}_T$ deviates from the intended Gaussian prior.
Crucially, the impact of this approximation depends on how the model is parameterized and trained, because these choices determine the schedules $a(t)$ and $b(t)$ and the conditioning of the inversion update. In particular, regimes in which the signal scale becomes small near $t=T$ amplify local errors through the division by $a(t)$ in Eq.~\ref{eq:general_ode_forward}, making violations of Eq.~\ref{eq:linear_assumption} more damaging.

\subsection{Parameterization}
\label{subsec:parametrization}
In this work we consider two representative parameterizations: DDIM, which is widely used in deterministic inversion and editing pipelines \cite{song2020denoisingdiffusionimplicitmodels}, and EDM, which is widely used for high-fidelity image generation \cite{karras2022elucidatingdesignspacediffusionbased}. All details on how the coefficients $a(t)$ and $b(t)$ are obtained are provided in Appendix~\ref{sec:appendix_background}.

\paragraph{DDIM.}
DDIM is specified through a cumulative noise schedule $\bar{\alpha}_t$ (Appendix~\ref{sec:appendix_background:ddim}). A common deterministic discretization yields
\begin{equation}
    a(t)=\sqrt{\frac{\bar{\alpha}_{t-1}}{\bar{\alpha}_t}},\quad
    b(t)=\sqrt{\frac{1}{\bar{\alpha}_{t-1}}-1}-\sqrt{\frac{1}{\bar{\alpha}_{t}}-1}.
\end{equation}
The network is trained to predict the added noise,
$F_\theta(\mathbf{x}_t,y,t)=\boldsymbol{\epsilon}_\theta(\mathbf{x}_t,y,t)$, with the denoising objective
$\theta^\star=\arg\min_\theta\,\mathbb{E}_{t,\mathbf{x}_0,\boldsymbol{\epsilon}}\big[\|\boldsymbol{\epsilon}-\boldsymbol{\epsilon}_\theta(\mathbf{x}_t,y,t)\|_2^2\big]$, where $\mathbf{x}_t=\sqrt{\bar{\alpha}_t}\,\mathbf{x}_0+\sqrt{1-\bar{\alpha}_t}\,\boldsymbol{\epsilon}$.

\paragraph{EDM.}
EDM is specified through a noise-level schedule $\sigma_t$ (Appendix~\ref{sec:appendix_background:edm}), leading to the simple coefficients
\begin{equation}
    a(t)=\frac{\sigma_{t-1}}{\sigma_t},\quad
    b(t)=\frac{\sigma_t-\sigma_{t-1}}{\sigma_t}.
\end{equation}
In contrast to DDIM, the network is trained to predict a denoised sample $F_\theta(\mathbf{x}_t,y,t)=\mathbf{x}_{0,\theta}(\mathbf{x}_t,y,t)$ under noisy inputs $\mathbf{x}_t=\mathbf{x}_0+\sigma_t\boldsymbol{\epsilon}$, using a weighted regression loss of the form
$\theta^\star=\arg\min_\theta\,\mathbb{E}_{t,\mathbf{x}_0,\boldsymbol{\epsilon}}\big[w(\sigma_t)\,\|F_\theta(\mathbf{x}_t,y,t)-\mathbf{x}_0\|_2^2\big]$.
An additional ingredient is \emph{preconditioning}: the network operates on a rescaled version of $\mathbf{x}_t$ so that it sees approximately unit-variance inputs for all $t$ \cite{karras2022elucidatingdesignspacediffusionbased}.
\section{Related Work}
\label{sec:related_work}

\paragraph{Deterministic Diffusion Inversion}
Recovering the latent noise $\mathbf{z}_T$ that maps back to a given image $\mathbf{x}_0$ with inversion is a cornerstone of modern editing workflows.
The standard approach leverages the PF-ODE \cite{song2021scorebasedgenerativemodelingstochastic}, which theoretically enables an exact bijective mapping between data and noise. DDIM Inversion \cite{song2020denoisingdiffusionimplicitmodels} approximates this trajectory, but as noted in recent analyses, it suffers from discretization errors due to the \textit{linear assumption} \cite{wallace2022edictexactdiffusioninversion}, leading to trajectory drift where the reconstructed image deviates from the input.
To address this, methods like Null-Class Inversion \cite{mokady2022nulltextinversioneditingreal}, ReNoise \cite{garibi2024renoiserealimageinversion}, and Negative-Prompt Inversion \cite{miyake2023negativepromptinversionfastimage} employ optimization to refine the conditioning or noise estimate, while EDICT \cite{wallace2022edictexactdiffusioninversion} or BDIA \cite{zhang2023exactdiffusioninversionbidirectional} enforces exactness via invertible affine coupling layers.

\paragraph{Stochastic Encoding and Cycle-Consistency}
Distinct from deterministic inversion, a parallel body of work leverages the forward stochastic process for encoding.
SDEdit \cite{meng2021sdeditguidedimagesynthesis} avoids explicit inversion entirely by injecting noise to an intermediate timestep $t$ (forward SDE) and sampling back. While effective for simple denoising, this "destructive" encoding lacks structural guidance, thus being prone to hallucinating content unrelated to the source input. 
CycleDiff \cite{zou2025cyclediffcyclediffusionmodels} formalizes this within a unified framework, treating the forward process as a stochastic encoder and enforcing cycle consistency to preserve semantics.
\textit{Distinction:} Unlike our approach, these methods do not aim to solve the deterministic reverse ODE for $\mathbf{z}_T$. Instead, they rely on stochastic noise injection to bridge domains. Our analysis specifically targets the limitations of the deterministic inversion path, and aims to correct the variance-structure trade-off directly in the ODE dynamics.

\paragraph{Spectral Bias in Generative Models}
Neural networks exhibit a well-documented "spectral bias," learning low-frequency variations significantly faster than high-frequency textures \cite{rahaman2019spectralbiasneuralnetworks}. This phenomenon has been extensively studied in the context of \textit{sampling}: generated images often suffer from "oversmoothing" when the model defaults to the conditional mean \cite{pfrommer2025conditionaldiffusionmodelactually, jiang2021focalfrequencylossimage, benita2025spectralanalysisdiffusionmodels}.
However, the manifestation of spectral bias during the \textit{inversion} process remains underexplored.
\citealt{staniszewski2025againrelationnoiseimage} recently pointed out a similar behavior, observing that inverted latents are less diverse and retain source patterns in "\textit{redundant pixel areas}". However, while they attribute this to errors in the early inversion steps, we trace the root cause to the $\boldsymbol{\epsilon}$-prediction objective itself.

\paragraph{Gradient Manipulation \& Orthogonal Projection}
Our proposed solution draws inspiration from gradient manipulation techniques used in multi-task learning, notably PCGrad \cite{yu2020gradientsurgerymultitasklearning}, which projects conflicting gradients onto the normal plane of other tasks to prevent interference. \\
We adapt this geometric intuition to the problem of variance injection: rather than projecting gradients to avoid task interference, we project variance correction to be orthogonal to the structural gradient of the source image. This allows us to restore the noise magnitude required for texture (solving the gaussianity gap) without corrupting the semantic layout (avoiding structural hallucination).

\section{Observations}
\label{sec:observations}

To understand the limitations of current inversion techniques for unpaired translation, we analyze the statistical properties of the inverted latents and their correlation with generation quality. We observe that the inverted noise maps systematically deviate from the theoretical Gaussian prior, exhibiting distinct spectral signatures that drive specific failure modes. To rigorously quantify these anomalies, we introduce specific metrics across the datasets studied (BBBC021 $\times8, \times16, \times32$ and Edges2Shoes). These metrics serve as the analytical backbone for the rest of this work.

\paragraph{Spectral Collapse: Noise Map \& Image Dependency}
\label{subsec:obs_spectral_collapse}
We first observe that when the source image is spectrally sparse (i.e. lacking significant high-frequency energy), standard deterministic inversion fails to fully whiten the signal. Consequently, the latent code $\mathbf{z}_T$ retains a visible low-frequency structural residual rather than converging to pure Gaussian noise.
\begin{figure}[!t]
    \centering
    \includegraphics[width=0.85\linewidth]{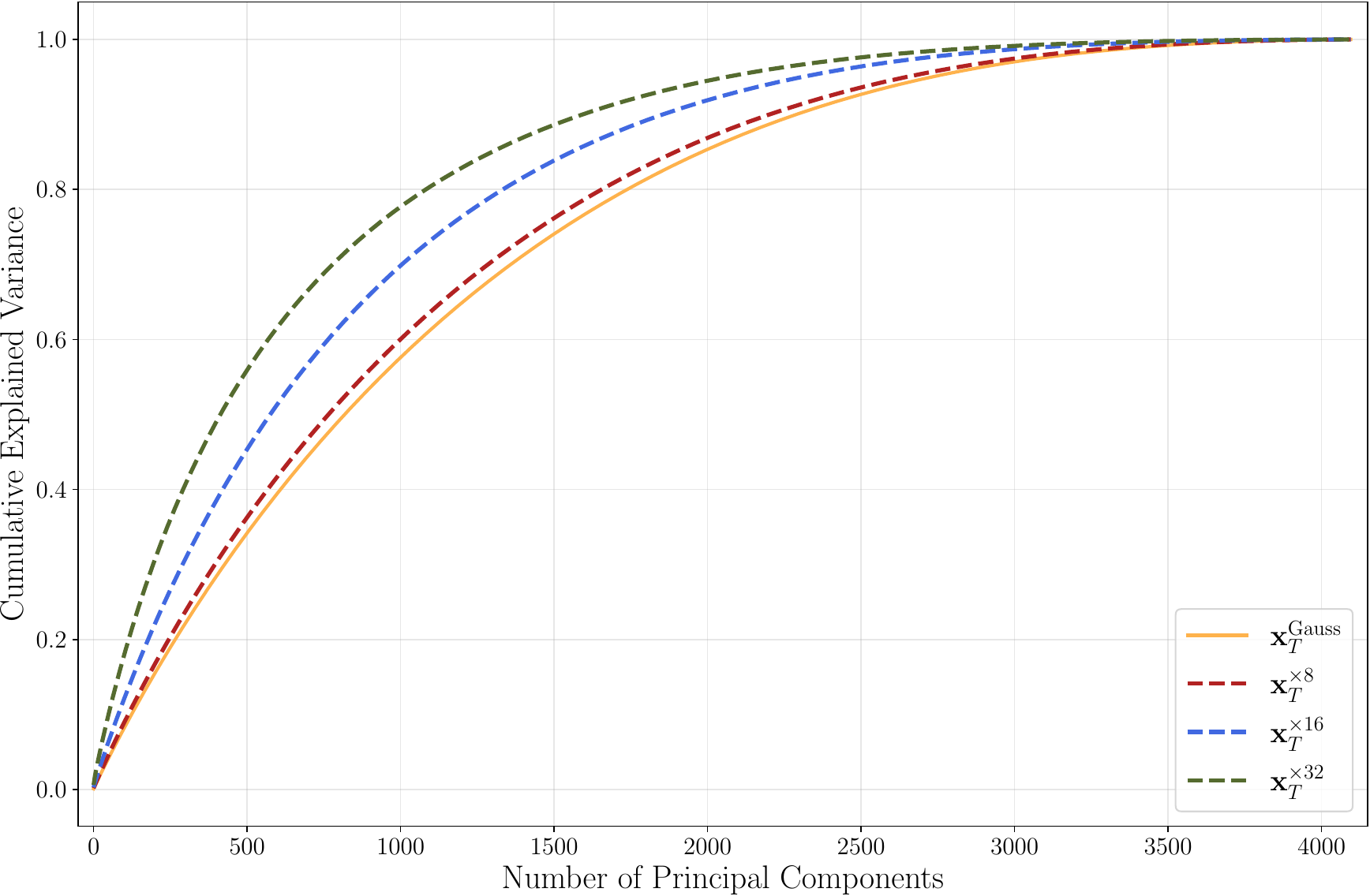}
    \caption{\textbf{Latent Entropy Analysis.}
    We plot the mean of Cumulative Explained Variance over 2000 noise maps via Principal Component Analysis (PCA).
    In contrast to isotropic Gaussian noise ($\mathbf{x}_T^{\mathrm{Gauss}}$) which uses all dimensions equally, noise maps ($\mathbf{x}_T^{\times{8}}$,$\mathbf{x}_T^{\times{16}}$, $\mathbf{x}_T^{\times{32}}$) exhibit dimensionality collapse, concentrating variance in a sparse set of components.}
    \label{fig:pca_entropy}
\end{figure}

\begin{figure}[ht!]
    \newcommand{\figheight}{3cm}
    \centering
    \subfloat[$ S_{\mathrm{HF}}$ vs. $ S_{\mathrm{DC}}$]{
        \includegraphics[height=\figheight]{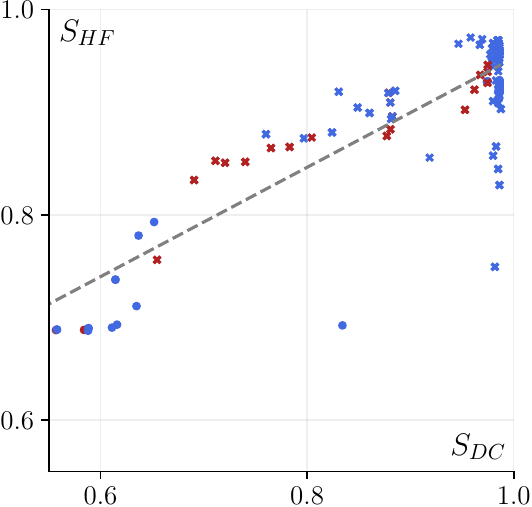}
        \label{fig:hf_vs_non_gaussiannity}
    }\hfill
    \subfloat[$ S_{\mathrm{LF}}$ vs. $ S_{\mathrm{DC}}$]{
        \includegraphics[height=\figheight]{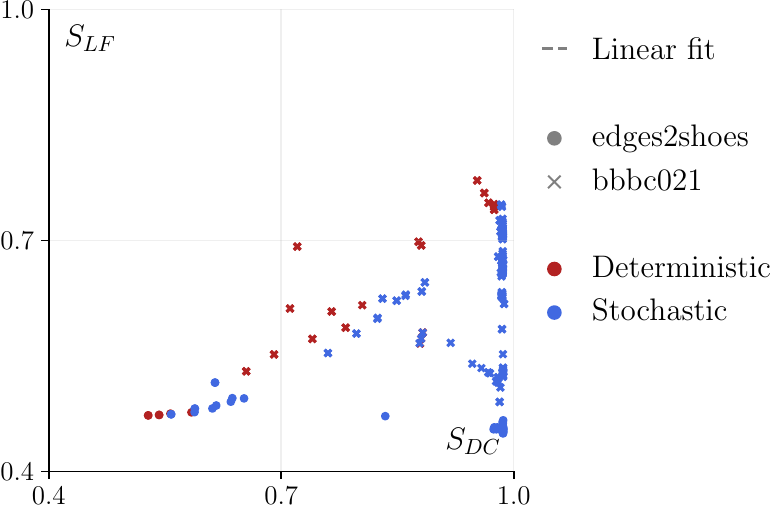}
        \label{fig:lf_vs_non_gaussiannity}
    }
    \caption{\textbf{High-Frequency \& Low-Frequency Scores vs. Correlation} We observe a strong positive correlation between $S_{\mathrm{HF}}$ and $S_{\mathrm{DC}}$, confirming that latent decorrelation is a prerequisite for high-frequency synthesis. In contrast, while non-Gaussian latents naturally preserve input structure (high $S_{\mathrm{LF}}$), enforcing Gaussianity via stochastic injection forces a trade-off that degrades structural fidelity.}
    \label{fig:hf_lf_vs_non_gaussiannity}
\end{figure}

\begin{itemize}[topsep=0pt, leftmargin=*]
    \item \textbf{Patch Correlation ($S_{\mathrm{DC}}$):} We observe strong spatial correlations between patches of the inverted latent. To quantify this, we introduce the \textbf{Decorrelation Score ($S_{\mathrm{DC}}$)}, which measures the dependence between spatial patches of the latent code. A high $S_{\mathrm{DC}}$ indicates that the latent is spatially structured (see Appendix~\ref{subsec:sg_patch}).
    \item \textbf{Gaussianity \& Entropy ($S_{\mathrm{G}}$):} This spatial structure correlates with a deviation from the theoretical isotropic Gaussian prior. We visualize this entropic gap in Figure~\ref{fig:pca_entropy}: inverted latents occupy a low-dimensional subspace compared to standard Gaussian noise, a phenomenon that intensifies as the input becomes spectrally sparser (e.g., $\times 32$ vs $\times 8$). We quantify this deviation using the \textbf{Gaussianity Score ($S_{\mathrm{G}}$)}, computed via a Kolmogorov-Smirnov test on the latent norms (see \ref{subsec:sg}). As shown in the correlation matrix, $S_{\mathrm{DC}}$ and $S_{\mathrm{G}}$ are strongly correlated (Fig.~\ref{fig:metric_correlation_matrix}).
    \item \textbf{High-Frequency Score ($S_{\mathrm{HF}}$):} We introduce the $S_{\mathrm{HF}}$ metric to assess the spectral fidelity of the generated images compared to the target domain. We observe a strong positive correlation between $S_{\mathrm{HF}}$ and the latent patch correlation metric $S_{\mathrm{DC}}$ (see Fig.~\ref{fig:hf_vs_non_gaussiannity} and Fig.~\ref{fig:metric_correlation_matrix}), confirming that structural artifacts in the latent directly degrade texture generation.
    \item \textbf{Impact on Generation:} Importantly, the non-Gaussianity of the latent acts as a bottleneck for texture generation. A latent with low variance (low $S_{\mathrm{G}}$) lacks the high-frequency energy required to seed realistic textures, leading to \textit{spectral collapse} in the final output. This causal link between latent statistics and image spectrum is visually demonstrated in Fig.~\ref{fig:spectral_collapse_psd} and quantitatively supported by the correlation analysis in Fig.~\ref{fig:metric_correlation_matrix}.
\end{itemize}

\begin{table}[!b]
    \centering
    \caption{Difference for each metric between \textit{deterministic} (standard, DirectInv, Null-Class) and \textit{stochastic} (TABA, ReNoise) on all datasets over 3 seeds.}
    \begin{tabular}{l | c c c c}
\toprule
\textbf{Statistic} & \textbf{$S_{HF}$} & \textbf{$S_{LF}$} & \textbf{$S_{DC}$} & \textbf{$S_{G}$} \\
\midrule
$\%\;\mathrm{improved}$ & 93.8\% & 37.5\% & 100.0\% & 75.0\% \\
\bottomrule
\end{tabular}
    \label{tab:ablations:stoch_vs_det}
\end{table}

\paragraph{Parameterization Trade-off}
\label{subsec:obs_tradeoff}
To address the issue of spectral collapse, we investigate potential off-the-shelf solutions and their limitations.
\begin{itemize}[topsep=0pt, leftmargin=*]
    \item \textbf{Impact of the Training Objective ($\boldsymbol{\epsilon}$ vs. $\mathbf{x}_0$):} We find that switching the prediction objective from $\boldsymbol{\epsilon}$-prediction to $\mathbf{x}_0$-prediction effectively eliminates spectral collapse in the noise map, regardless of the specific sampling framework (e.g., DDIM with $\mathbf{x}_0$-pred or EDM, see Table.~\ref{tab:ablations:prediction_target}). For our experiments, we adopt EDM \cite{karras2022elucidatingdesignspacediffusionbased} as the representative $\mathbf{x}_0$-model due to its superior training stability and state-of-the-art performances.

    \item \textbf{Structure-Texture Tradeoff:} However, this spectral improvement comes at a cost. To quantify this, we introduce the \textbf{Low-Frequency Score ($S_{\mathrm{LF}}$)}, which assesses the structural alignment between the generated output and the input condition (see Appendix~\ref{subsec:slf}). We find that while $\mathbf{x}_0$-prediction (e.g., EDM) successfully restores high-frequency texture (high $S_{\mathrm{HF}}$), it suffers from \textit{structural drift}, yielding lower $S_{\mathrm{LF}}$ scores than $\boldsymbol{\epsilon}$-prediction. This motivates the need for a hybrid approach that still enforces structural guidance within a spectrally robust framework.
\end{itemize}

\section{Proposed Method}

\subsection{Orthogonal Control Drift Terms}
\label{subsec:orthogonal_drifts}
Based on previous observations, our goal is twofold:
(i) \emph{enforce Gaussianity} along the inversion trajectory to recover high-frequency realism, and
(ii) \emph{preserve structure} by explicitly correcting low-frequency deviations.

At any time step $t$, given $\mathbf{x}_t$, the diffusion model provides both a noise estimate $\boldsymbol{\epsilon}_\theta(\mathbf{x}_t,y,t)$ and a reconstruction $\hat{\mathbf{x}}_{0,\theta}(\mathbf{x}_t,y,t)$ (see Sec.~\ref{sec:background}).
We leverage these predictions to build two losses and corresponding correction directions.

\paragraph{Enforcing Gaussianity (high-frequency correction)}
We penalize deviations of the predicted noise energy from its theoretical expectation.
We define:
\begin{align}
    L_{\mathrm{HF}}(\mathbf{x}_t)               & \coloneqq \big(\|\boldsymbol{\epsilon}_\theta(\mathbf{x}_t,y,t)\|_2^2 - d\big)^2, \\
    \mathrm{and} \quad \mathbf{g}_{\mathrm{HF}} & \coloneq \nabla_{\mathbf{x}_t} L_{\mathrm{HF}}(\mathbf{x}_t). \label{eq:lhf_def}
\end{align}
Minimizing this loss leverages the concentration of measure in high-dimensional spaces. Since a Gaussian vector $\mathbf{z} \in \mathbb{R}^d$ concentrates tightly on the hypersphere of radius $\sqrt{d}$ (where $\mathbb{E}[\|\mathbf{z}\|^2] = d$), $L_{\mathrm{HF}}$ effectively constrains the latent to the correct energy shell. This prevents the variance collapse that leads to texture loss, serving as a robust proxy for restoring Gaussian statistics.

\paragraph{Rectifying low-frequency perturbations (structure correction).}
To preserve the conditioning content (e.g., low-resolution observation, sketch constraints), we penalize reconstruction drift in the denoised estimate.
Let $\mathbf{x}_0^{\mathrm{LR}}$ denote the low-frequency reference derived from the conditioning input.
We define:
\begin{align}
    L_{\mathrm{LF}}(\mathbf{x}_t)               & \coloneq
    \big\|\hat{\mathbf{x}}_{0,\theta}(\mathbf{x}_t,y,t) - \mathbf{x}_0^{\mathrm{LR}}\big\|_2^2,                                    \\
    \mathrm{and} \quad \mathbf{g}_{\mathrm{LF}} & \coloneq \nabla_{\mathbf{x}_t} L_{\mathrm{LF}}(\mathbf{x}_t). \label{eq:llf_def}
\end{align}
Note that $\hat{\mathbf{x}}_{0,\theta}$ is derived from the model output. For example, in DDIM-style parameterization, $\hat{\mathbf{x}}_{0,\theta}(\mathbf{x}_t) = (\mathbf{x}_t - \sqrt{1-\bar{\alpha}_t}\boldsymbol{\epsilon}_\theta(\mathbf{x}_t)) / \sqrt{\bar{\alpha}_t}$.

\paragraph{Orthogonal drifts.}
Inspired by gradient projection techniques in multi-task learning (e.g., PCGrad \cite{yu2020gradientsurgerymultitasklearning}), we apply orthogonal projection specifically when the objectives are \emph{conflicting} (i.e., when their cosine similarity is negative, $\langle \mathbf{g}_{\mathrm{HF}}, \mathbf{g}_{\mathrm{LF}} \rangle < 0$).
In such cases, we project each update onto the normal plane of the other to resolve interference:
\begin{align}
    \mathbf{u}_{\mathrm{HF}}(\mathbf{x}_t)
     & = -\eta_{\mathrm{HF}}\left( \mathbf{g}_{\mathrm{HF}} - \frac{\langle \mathbf{g}_{\mathrm{HF}},\mathbf{g}_{\mathrm{LF}}\rangle}{\|\mathbf{g}_{\mathrm{LF}}\|_2^2}\,\mathbf{g}_{\mathrm{LF}} \right),
    \label{eq:u_hf}
    \\
    \mathbf{u}_{\mathrm{LF}}(\mathbf{x}_t)
     & = -\eta_{\mathrm{LF}}\left( \mathbf{g}_{\mathrm{LF}} - \frac{\langle \mathbf{g}_{\mathrm{LF}},\mathbf{g}_{\mathrm{HF}}\rangle}{\|\mathbf{g}_{\mathrm{HF}}\|_2^2}\,\mathbf{g}_{\mathrm{HF}} \right).
    \label{eq:u_lf}
\end{align}
If the gradients agree ($\langle \mathbf{g}_{\mathrm{HF}}, \mathbf{g}_{\mathrm{LF}} \rangle \geq 0$), we retain the standard updates.
We perform this selective projection because the noise estimate $\boldsymbol{\epsilon}_\theta$ and the clean data estimate $\hat{\mathbf{x}}_{0,\theta}$ are \textbf{linearly coupled} via $\mathbf{x}_t$. When the objectives pull in opposite directions, a naive variance update could degrade structure. Our approach ensures that, in these conflict scenarios, high-frequency energy is injected strictly into the null-space of the structural gradient.

\begin{figure}[ht!]
    \centering
    \includegraphics[width=1.0\columnwidth]{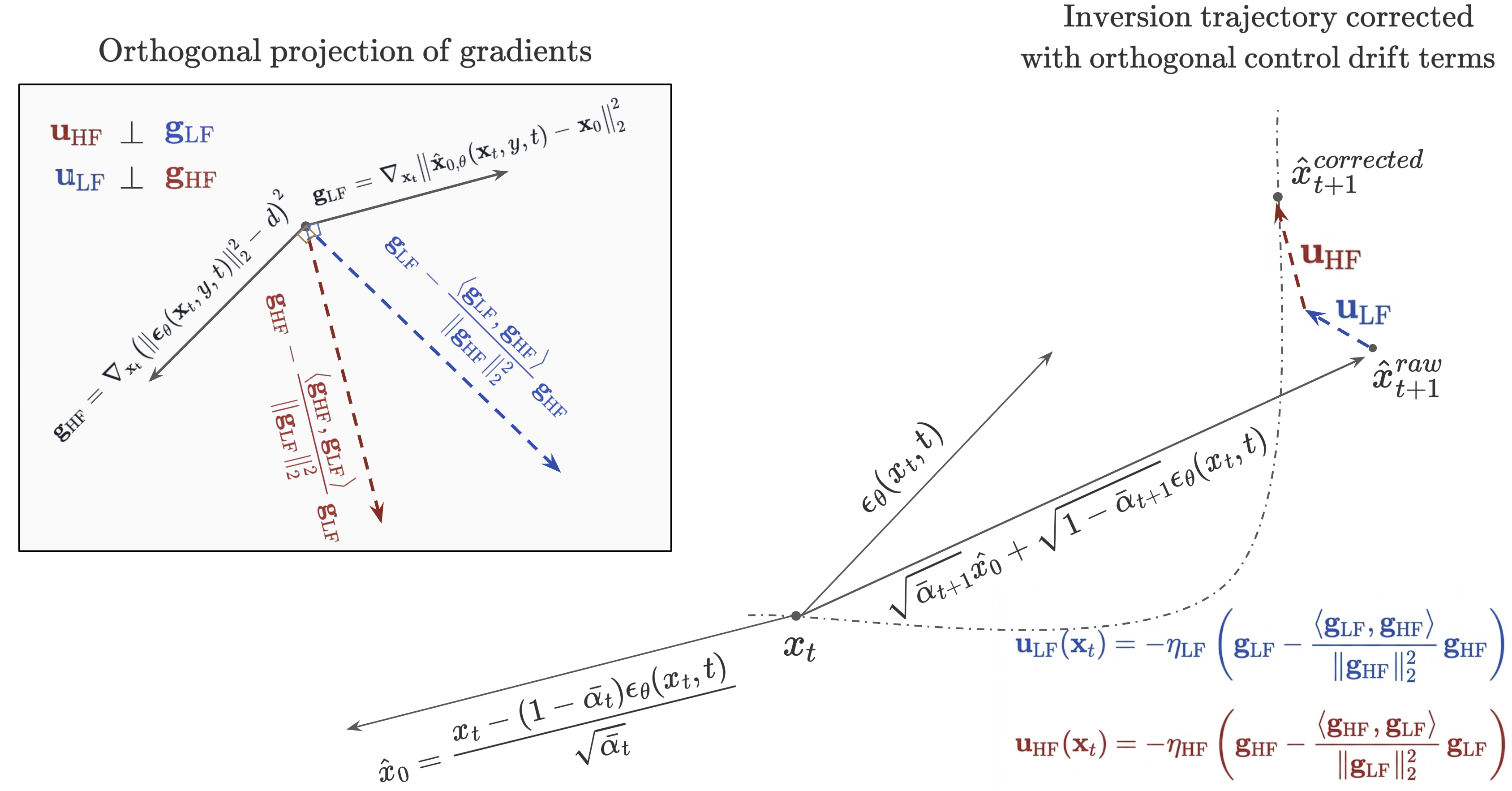}
    \caption{\textbf{Inversion trajectory corrected with Orthogonal Variance Guidance} At each timestep $t$, we compute gradients for variance ($\mathbf{g}_{\mathrm{HF}}$) and structure ($\mathbf{g}_{\mathrm{LF}}$) and project them into mutually orthogonal subspaces if conflicting to guide the ODE trajectory.}
    \label{fig:inversion_trajectory}
\end{figure}

\paragraph{Corrected inversion update.}
Let the base deterministic inversion step (PF-ODE discretization) be:
\begin{equation}
    \mathbf{x}_{t+1}^{\mathrm{base}} = \frac{\mathbf{x}_t - b(t)\,F_\theta(\mathbf{x}_t,\bar{y},t)}{a(t)}.
    \label{eq:general_ode_forward_base}
\end{equation}
We propose to augment this update with the orthogonal drifts computed at $\mathbf{x}_t$:
\begin{equation}
    \mathbf{x}_{t+1}
    = \mathbf{x}_{t+1}^{\mathrm{base}} + \mathbf{u}_{\mathrm{LF}}(\mathbf{x}_t) + \mathbf{u}_{\mathrm{HF}}(\mathbf{x}_t).
    \label{eq:general_ode_forward_corrected}
\end{equation}
Equivalently, one may view Eq.~\eqref{eq:general_ode_forward_corrected} as integrating the PF-ODE with an additional control drift decomposed into two non-interfering orthogonal components.
\Cref{fig:inversion_trajectory} illustrates how these terms correct the inversion trajectory.

\subsection{Observing how $\eta_{\mathrm{LF}}$ and $\eta_{\mathrm{HF}}$ affect performance}
\label{subsec:eta_tradeoff}
\begin{figure}[!b]
    \centering
    \includegraphics[width=0.9\columnwidth]{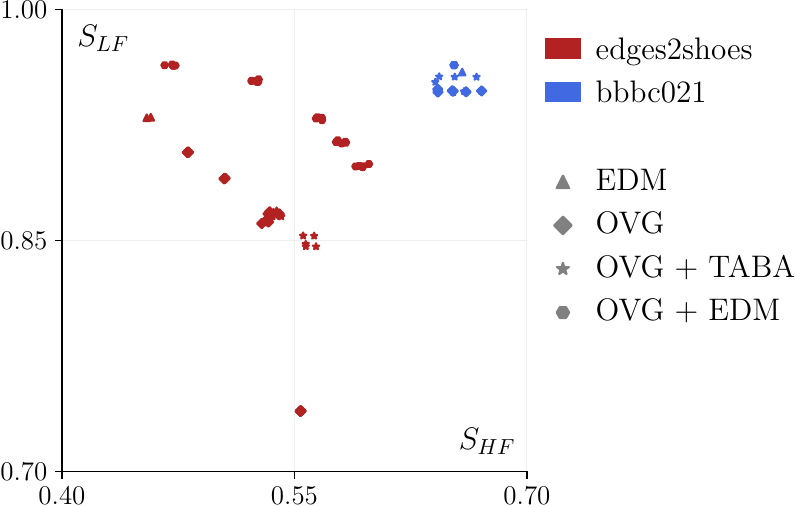}
    \caption{\textbf{Performance Pareto Frontier ($S_{\mathrm{LF}}$ vs. $S_{\mathrm{HF}}$).}
        We plot the performance landscape obtained by sweeping guidance scales $(\eta_{\mathrm{HF}}, \eta_{\mathrm{LF}})$ across datasets.
        Unlike baselines constrained to a lower performance envelope, OVG expands the Pareto frontier, enabling the selection of operating points that simultaneously maximize structural fidelity and high-frequency texture.}
    \label{fig:pareto_tradeoff}
\end{figure}

While the orthogonal projection mitigates the directional conflict between gradients, the step sizes $\eta_{\mathrm{HF}}$ and $\eta_{\mathrm{LF}}$ allow us to calibrate the \textbf{magnitude} of the guidance to the specific dataset (e.g., spectrally sparser inputs may require stronger $\eta_{\mathrm{HF}}$).
As seen in Figure~\ref{fig:pareto_tradeoff}, this decoupling allows OVG to expand the Pareto frontier of the model: unlike baseline methods that sacrifice structure for texture (or vice-versa), our method allows us to select a point where both metrics are maximized simultaneously.
In practice, we determine the optimal $(\eta_{\mathrm{HF}},\eta_{\mathrm{LF}})$ configuration via a grid search, selecting the values that yield the best joint performance.
\section{Experiments}
\label{sec:experiments}

\subsection{Experimental Setup}

\paragraph{Datasets.} We evaluate our method across distinct modalities to test robustness on \textit{asymmetric entropy} domains--tasks where the source domain is spectrally sparse (i.e, information-poor) compared to the spectrally dense target.
\begin{itemize}[topsep=0pt, leftmargin=*]
    \item \textbf{Edges2Shoes} \cite{isola2017image}: Standard edge-to-shoe benchmark. We use 49k strictly unpaired images to simulate realistic domain adaptation, requiring complex texture synthesis from sparse semantic layouts.
    \item \textbf{BBBC021} \cite{caie2010high}: Biological microscopy dataset adapted for super-resolution. We rigorously test spectral collapse via aggressive nearest-neighbor downsampling ($\times8, \times16, \times32$) in a strictly unpaired setting, where unique samples appear exclusively in either high or low resolution.
\end{itemize}

\paragraph{Baselines.} We compare against state-of-the-art inversion-based editing methods, categorized by their strategy for handling the latent noise trajectory:
\begin{itemize}[topsep=0pt, leftmargin=*]
    \item \textbf{Deterministic Inversion:} We evaluate standard DDIM Inversion \cite{song2020denoisingdiffusionimplicitmodels} (naive ODE reversal), Null-Class Inversion \cite{mokady2022nulltextinversioneditingreal} which optimizes the unconditional embedding to minimize reconstruction error, and DirectInversion \cite{ju2023directinversionboostingdiffusionbased} which mitigates drift from $\mathbf{x}_0$ by separating source and target trajectories.
    \item \textbf{Stochastic/Noise-Injection:} We compare against ReNoise \cite{garibi2024renoiserealimageinversion}, which applies iterative corrections to the noise maps to recover lost details, and ``There and Back Again'' (TABA) \cite{staniszewski2025againrelationnoiseimage}, which replaces early inversion steps with stochastic forward diffusion to better approximate the Gaussian prior.
    \item \textbf{Alternative Parameterization (EDM \& $\mathbf{x}_0$-Ablation):} We employ a conditional EDM \cite{karras2022elucidatingdesignspacediffusionbased} model as a state-of-the-art reference for $\mathbf{x}_0$-prediction.
          To strictly isolate the impact of the prediction target from EDM's specific design choices (e.g., noise schedule, preconditioning), we additionally train a DDIM with $\mathbf{x}_0$-prediction (see \Cref{sec:experiments:epsilon_vs_x0}).
          \textit{Note:} We do not combine EDM formulation with TABA, since EDM's boundary condition ($\sigma \ge \sigma_{\min} \ge 0$) \textit{intrinsically} resolves singularity that TABA targets through, effectively performing the necessary stochastic injection by design.
\end{itemize}

\begin{table*}[!t]
    \centering
    \caption{\textbf{Quantitative Comparison.} We evaluate translation quality across 2 representative datasets. We report \textbf{PSNR} ($\uparrow$), \textbf{MS-SSIM} ($\uparrow$), \textbf{LPIPS} ($\downarrow$), and \textbf{FID} ($\downarrow$) for general quality; \textbf{HF Score} ($\uparrow$) and \textbf{LF Score} ($\uparrow$) for spectral fidelity; and \textbf{Decorrelation Score} ($\uparrow$) to measure spectral collapse. All values are mean over 3 seeds. Bold is best and underline is within 95\% of the best.}
    \label{tab:main_results}
    \resizebox{\textwidth}{!}{
        \setlength{\tabcolsep}{3pt} 
                \begin{tabular}{l | c c c c c c c | | c c c c c c c}
            \toprule
             & \multicolumn{7}{c|}{\textbf{Edges2Shoes}} & \multicolumn{7}{c}{\textbf{BBBC021 ($\times$16)}} \\
            \textbf{Method} & \textbf{PSNR} & \textbf{MS-SSIM} & \textbf{LPIPS} & \textbf{FID} & \textbf{$S_{HF}$} & \textbf{$S_{LF}$} & \textbf{$S_{DC}$} & \textbf{PSNR} & \textbf{MS-SSIM} & \textbf{LPIPS} & \textbf{FID} & \textbf{$S_{HF}$} & \textbf{$S_{LF}$} & \textbf{$S_{DC}$} \\
              & $\uparrow$ & $\uparrow$ & $\downarrow$ & $\downarrow$ & $\uparrow$ & $\uparrow$ & $\uparrow$ & $\uparrow$ & $\uparrow$ & $\downarrow$ & $\downarrow$ & $\uparrow$ & $\uparrow$ & $\uparrow$ \\
            \midrule
            \textit{Deterministic} &  &  &  &  &  &  &  &  &  &  &  &  &  &  \\
            Standard & 7.42 & 0.404 & 0.467 & 385.01 & 0.69 & 0.47 & 0.53 & 15.11 & 0.553 & 0.521 & 217.09 & 0.83 & 0.55 & 0.69 \\
            Null Class & 7.41 & 0.403 & 0.466 & 392.73 & 0.69 & 0.47 & 0.54 & \underline{16.22} & 0.616 & 0.434 & 186.78 & 0.85 & 0.61 & 0.71 \\
            Direct & 7.41 & 0.403 & 0.466 & 392.82 & 0.69 & 0.47 & 0.54 & \underline{16.22} & 0.616 & 0.434 & 186.76 & 0.85 & 0.61 & 0.71 \\
            \midrule
            \textit{Stochastic} &  &  &  &  &  &  &  &  &  &  &  &  &  &  \\
            Renoise & 7.95 & \underline{0.423} & 0.462 & 305.31 & 0.74 & 0.52 & 0.61 & \underline{16.20} & 0.606 & 0.457 & 145.57 & \underline{0.92} & 0.62 & 0.83 \\
            TABA & 8.79 & 0.398 & 0.463 & 93.32 & \underline{0.92} & 0.46 & \textbf{0.99} & 15.90 & 0.600 & 0.329 & \textbf{12.06} & \textbf{0.96} & 0.63 & \textbf{0.98} \\
            \midrule
            \textit{Ours} &  &  &  &  &  &  &  &  &  &  &  &  &  &  \\
            EDM & \textbf{9.59} & 0.420 & 0.440 & \textbf{53.10} & \textbf{0.93} & 0.46 & \underline{0.97} & \textbf{16.92} & \underline{0.634} & 0.294 & 14.21 & \textbf{0.96} & \underline{0.66} & \textbf{0.98} \\
            OVG & 8.58 & \underline{0.427} & 0.433 & 145.86 & 0.87 & 0.54 & \underline{0.98} & \underline{16.66} & \textbf{0.665} & \underline{0.281} & 18.62 & \underline{0.95} & \textbf{0.67} & \underline{0.97} \\
            OVG + TABA & 8.49 & \underline{0.436} & 0.424 & 162.59 & 0.85 & 0.56 & \underline{0.98} & \underline{16.54} & \underline{0.656} & \textbf{0.279} & 13.14 & \textbf{0.96} & \textbf{0.67} & \textbf{0.98} \\
            OVG + EDM & \underline{9.38} & \textbf{0.443} & \textbf{0.381} & 90.16 & \underline{0.90} & \textbf{0.60} & 0.90 & \underline{16.90} & 0.621 & 0.299 & 17.37 & \textbf{0.96} & \underline{0.65} & \textbf{0.98} \\
            \bottomrule
        \end{tabular}}
\end{table*}
\paragraph{Metrics.} To quantitatively assess the trade-off between textural \textit{realism} and structural \textit{fidelity}, we employ a diverse suite of metrics (detailed in Appendix \ref{sec:appendix_metrics}). In each table, we present the best metric in bold, and if a metric is within 95\% of the best value it is underlined.
\begin{itemize}[topsep=0pt, leftmargin=*]
    \item \textbf{Standard Perceptual Metrics (FID, LPIPS, MS-SSIM, HaarPSI):} We adhere to standard protocols for domain adaptation. We report LPIPS and MS-SSIM to quantify perceptual quality. We also use Fréchet Inception Distance \cite{heusel2017ganstrainedtimescaleupdate} to assess the semantic validity of the generated distribution. This is essential to complement our spectral metrics: while $S_{\mathrm{HF}}$ measures energy magnitude, FID ensures that this high-frequency energy manifests as realistic texture (e.g., leather grain) rather than unstructured noise. To ensure statistical robustness, all results are reported as the mean over 3 random seeds. We employ FID  as a perceptual proxy for texture realism, to verify spectral recovery aligns with semantic plausibility rather than as strict large-scale distributional metric.

    \item \textbf{Spectral \& Structural Diagnostics ($S_{\mathrm{HF}}$, $S_{\mathrm{LF}}$, $S_{\mathrm{DC}}$):} Leveraging the spectral suite defined in Sec.~\ref{sec:observations}, we explicitly quantify the structure-texture trade-off. We track $S_{\mathrm{HF}}$ to measure spectral energy recovery (preventing collapse) and $S_{\mathrm{LF}}$ to monitor input fidelity (preventing drift), while $S_{\mathrm{DC}}$ serves as a diagnostic for spatial artifacts in the latent space. Detailed formal definitions and implementation specifics are provided in Appendix~\ref{sec:appendix_metrics}.

    \item \textbf{Metric Validation:} We validate these spectral metrics by analyzing their cross-correlation with other standard scores. As shown in the \Cref{fig:metric_correlation_matrix}, our spectral metrics align strongly with human-perceived quality while providing the granular separation of frequency bands required to analyze the structure-texture trade-off.
\end{itemize}

\paragraph{Implementation Details.}
We implement our primary framework using a standard conditional U-Net backbone (SD v1.5 \cite{rombach2022highresolutionimagesynthesislatent}).
To verify architectural robustness, we also validate performance using a DiT-L/2 \cite{peebles2023scalablediffusionmodelstransformers} in our ablation studies.
All models are trained using AdamW optimization.
To account for the distinct spectral properties of each modality (Pixel vs. Latent) and task complexity, we calibrate training hyperparameters - including batch size, learning rate schedules, and EMA decay rates—specifically for each dataset.
Detailed configurations for all experiments, including the exact EDM preconditioning parameters ($\sigma_{\text{data}}, \sigma_{\min}, \sigma_{\max}$), are provided in \textbf{Appendix~\ref{sec:appendix_training_details}}.

\subsection{Main Results: Unpaired Translation Quality}
Table \ref{tab:main_results} summarizes the performance across all datasets.

\paragraph{Texture Recovery vs. Latent Statistics.}
Consistent with our hypothesis, deterministic methods (Standard, Null-Class) exhibit low $S_{\mathrm{DC}}$, confirming that the latents retain strong spatial correlations. This correlates directly with their low HF Scores ($S_{\mathrm{HF}} \approx 0.69$ on Edges2Shoes), resulting in oversmoothed outputs. Stochastic methods such as TABA successfully restore latent independence ($S_{\mathrm{DC}} \to 0.99$) and high-frequency texture ($S_{\mathrm{HF}} \ge 0.92$). However, this comes at the cost of structural fidelity, as indicated by lower $S_{\mathrm{LF}}$ scores compared to our optimal configurations (e.g., 0.46 vs 0.60 on Edges2Shoes). We provide  quantitative comparison in \Cref{tab:ablations:stoch_vs_det} and qualitative comparisons visually confirming these findings in \Cref{fig:comparison_methods_grid} and Appendix~\ref{sec:appendix_qualitative}.

\paragraph{OVG Performance.}
Our method achieves the state-of-the-art trade-off between realism and fidelity. Crucially, on the challenging Edges2Shoes dataset, the {EDM+OVG} configuration achieves the best perceptual quality (LPIPS 0.381) while significantly outperforming TABA in structural preservation ($S_{\mathrm{LF}}$ 0.60 vs 0.46). This confirms that by injecting variance orthogonally, OVG recovers texture without disrupting the semantic layout of the input.

\subsection{Ablation Studies and Analysis}

\paragraph{Impact of Backbone Architecture (U-Net vs. DiT).}
To investigate if spectral collapse is an artifact of convolution layers, we replicated the BBBC021 experiments using a Diffusion Transformer (DiT) backbone.
As shown in Table \ref{tab:ablations:arch}, both U-Net and DiT exhibit identical spectral collapse patterns under standard inversion ($S_{HF} <0.89$, $S_{DC} < 0.89$), and both recover texture similarly under OVG ($S_{HF} = 0.96$, $S_{DC} = 98$). This effectively rules out the inductive bias of convolutional layers as the primary cause, suggesting instead that the smoothness bias is intrinsic to the \textit{inversion dynamics} itself.

\begin{table}[H]
    \centering
    \caption{Ablations on Backbone Architecture for BBBC021 $\times$16.}
    \resizebox{\columnwidth}{!}{%
        \begin{tabular}{l | l | c c c c c c c}
\toprule
\textbf{Arch.} & \textbf{Method} & \textbf{PSNR} & \textbf{MS-SSIM} & \textbf{LPIPS} & \textbf{FID} & \textbf{$S_{HF}$} & \textbf{$S_{LF}$} & \textbf{$S_{DC}$} \\
\midrule
\multirow{3}{*}{U-Net} & Standard & 15.11 & 0.553 & 0.521 & 217.09 & 0.83 & 0.55 & 0.69 \\
 & TABA & \underline{15.90} & 0.600 & 0.329 & \textbf{12.06} & \textbf{0.96} & 0.63 & \textbf{0.98} \\
 & OVG & \textbf{16.54} & \textbf{0.656} & \textbf{0.279} & 13.14 & \textbf{0.96} & \textbf{0.67} & \textbf{0.98} \\
\midrule
\multirow{3}{*}{DiT} & Standard & \textbf{18.02} & \textbf{0.697} & 0.343 & 134.48 & 0.88 & \textbf{0.70} & 0.88 \\
 & TABA & \underline{17.55} & \underline{0.663} & 0.275 & 18.55 & \underline{0.95} & \underline{0.69} & \textbf{0.99} \\
 & OVG & \underline{17.77} & \underline{0.664} & \textbf{0.261} & \textbf{17.53} & \textbf{0.97} & \underline{0.69} & \underline{0.98} \\
\bottomrule
\end{tabular}
    }
    \label{tab:ablations:arch}
\end{table}

\paragraph{Impact of Diffusion Space (Latent vs. Pixel).}

We conducted a controlled experiment in pixel space, eliminating the autoencoder bottleneck while strictly maintaining all other experimental variables constant (DDIM framework, $\boldsymbol{\epsilon}$-prediction, 50 steps). We observe that the inversion dynamics remain unchanged: the pixel-space noise maps exhibit the same spectral collapse effect, displaying extreme spatial correlations $S_{DC}$ as their latent counterparts. This rules out VAE compression as the root cause, confirming that the smoothness bias is intrinsic to the deterministic inversion process itself.

\begin{table}[h]
    \centering
    \caption{Ablations on Dataspace for BBBC021 $\times$16.}
    \resizebox{\columnwidth}{!}{%
        \begin{tabular}{l | l | c c c c c c c}
\toprule
\textbf{Data} & \textbf{Method} & \textbf{PSNR} & \textbf{MS-SSIM} & \textbf{LPIPS} & \textbf{FID} & \textbf{$S_{HF}$} & \textbf{$S_{LF}$} & \textbf{$S_{DC}$} \\
\midrule
\multirow{3}{*}{Latent} & Standard & 15.11 & 0.553 & 0.521 & 217.09 & 0.83 & 0.55 & 0.69 \\
 & TABA & \underline{15.90} & 0.600 & 0.329 & \textbf{12.06} & \textbf{0.96} & 0.63 & \textbf{0.98} \\
 & OVG & \textbf{16.54} & \textbf{0.656} & \textbf{0.279} & 13.14 & \textbf{0.96} & \textbf{0.67} & \textbf{0.98} \\
\midrule
\multirow{3}{*}{Pixel} & Standard & 11.91 & 0.509 & 0.419 & 208.00 & \textbf{0.97} & 0.18 & 0.07 \\
 & TABA & \textbf{18.69} & \textbf{0.719} & \textbf{0.298} & \underline{105.34} & 0.87 & \textbf{0.73} & \textbf{0.87} \\
 & OVG & \textbf{18.69} & \underline{0.718} & \textbf{0.298} & \textbf{105.19} & 0.87 & \textbf{0.73} & \textbf{0.87} \\
\bottomrule
\end{tabular}
    }
    \label{tab:ablations:dataspace}
\end{table}

\paragraph{Robustness to Spectral Sparsity ($\times8$ to $\times32$).}
We stress-test the methods by increasing the downsampling factor on BBBC021. While OVG consistently offers the best perception-distortion trade-off at moderate sparsity ($\times8, \times16$), the extreme $\times32$ regime reveals distinct failure modes (Table~\ref{tab:ablations:resolution}). Standard inversion minimizes distortion (highest PSNR) but fails perceptually (FID 99.14), collapsing to a mean-prediction state lacking texture. Conversely, OVG and TABA methods maintain high realism (FID $\approx$ 15, $S_{\mathrm{HF}} \ge 0.95$), proving capable of synthesizing plausible biological textures even when the input signal is absent.

\begin{table}[H]
    \centering
    \caption{Ablations on Scale Factor for BBBC021.}
    \resizebox{\columnwidth}{!}{%
        \begin{tabular}{l | l | c c c c c c c}
\toprule
\textbf{Res.} & \textbf{Method} & \textbf{PSNR} & \textbf{MS-SSIM} & \textbf{LPIPS} & \textbf{FID} & \textbf{$S_{HF}$} & \textbf{$S_{LF}$} & \textbf{$S_{DC}$} \\
\midrule
\multirow{3}{*}{x8} & Standard & \underline{18.08} & \underline{0.733} & 0.269 & 32.41 & \underline{0.93} & \underline{0.75} & \underline{0.97} \\
 & TABA & \underline{18.05} & 0.724 & 0.256 & 17.00 & \underline{0.94} & \underline{0.74} & \textbf{0.98} \\
 & OVG & \textbf{18.92} & \textbf{0.770} & \textbf{0.194} & \textbf{15.42} & \textbf{0.96} & \textbf{0.78} & \textbf{0.98} \\
\midrule
\multirow{3}{*}{x16} & Standard & 15.11 & 0.553 & 0.521 & 217.09 & 0.83 & 0.55 & 0.69 \\
 & TABA & \textbf{18.69} & \textbf{0.719} & 0.298 & 105.34 & 0.87 & \textbf{0.73} & 0.87 \\
 & OVG & 16.54 & 0.656 & \textbf{0.279} & \textbf{13.14} & \textbf{0.96} & 0.67 & \textbf{0.98} \\
\midrule
\multirow{3}{*}{x32} & Standard & \textbf{14.58} & \textbf{0.529} & 0.459 & 99.14 & 0.90 & \textbf{0.58} & 0.88 \\
 & TABA & \underline{14.20} & 0.495 & \textbf{0.422} & \textbf{14.41} & \underline{0.95} & 0.55 & \textbf{0.99} \\
 & OVG & \underline{13.89} & 0.484 & \underline{0.423} & 15.34 & \textbf{0.97} & 0.53 & \underline{0.97} \\
\bottomrule
\end{tabular}}
    \label{tab:ablations:resolution}
\end{table}

\paragraph{Epsilon vs. $\mathbf{x}_0$-Prediction.}
\label{sec:experiments:epsilon_vs_x0}
To isolate the impact of the prediction target, we compare identical inversion configurations using either $\boldsymbol{\epsilon}$ or $\mathbf{x}_0$ parameterization across all scales and seeds (Table~\ref{tab:ablations:prediction_target}).
Results indicate that switching to $\mathbf{x}_0$-prediction is a definitive cure for spectral collapse: it improves high-frequency recovery ($S_{\mathrm{HF}}$) in 96.6\% of cases and latent decorrelation ($S_{\mathrm{DC}}$) in 89.7\% of cases.
However, this spectral restoration comes at a cost: structural fidelity ($S_{\mathrm{LF}}$) and reconstruction accuracy (PSNR, MS-SSIM) improve in only $\approx$ 17-35\% of trials.
This confirms that while $\mathbf{x}_0$-prediction solves the information bottleneck, it introduces \textit{structural drift} by hallucinating unconstrained texture, necessitating the targeted guidance of OVG to restore semantic fidelity. We provide theoretical investigation detailing the signal-to-noise dynamics driving this trade-off in Appendix~\ref{sec:appendix_theory}.

\begin{table}[H]
    \centering
    \caption{Ablations on Prediction Target for BBBC021. We plot the difference $\Delta$ between each metrics for $\boldsymbol{\epsilon}$-prediction and $\mathbf{x}_0$-prediction for the same inversion algorithm (standard, TABA or OVG) on all factor scale over 3 seeds.}
    \resizebox{\columnwidth}{!}{%
        \begin{tabular}{l | c c c c c c c}
\toprule
\textbf{Statistic} & \textbf{PSNR} & \textbf{MS-SSIM} & \textbf{LPIPS} & \textbf{FID} & \textbf{$S_{HF}$} & \textbf{$S_{LF}$} & \textbf{$S_{DC}$} \\
\midrule
$\%\;\mathrm{improved}$ & 79.3\% & 72.4\% & 96.6\% & 34.5\% & 96.6\% & 72.4\% & 89.7\% \\
\bottomrule
\end{tabular}
    }
    \label{tab:ablations:prediction_target}
\end{table}
\section{Discussion}
\label{sec:conclusion}
We identify \textit{spectral collapse} as a fundamental failure mode of deterministic inversion, driven by the breakdown of the solver's linear assumption in low-SNR regimes. This phenomenon is universal across architectures (U-Net, DiT) and domains (pixel, latent), confirming that the observed "smoothness bias" stems from inference dynamics rather than architectural priors. By enforcing a strict bijection, deterministic methods inherently preserve the spectral voids of the input (e.g., missing textures). We propose breaking this symmetry via Orthogonal Variance Guidance (OVG), which actively restores the latent variance (noise norm) while restricting updates to the structural null-space, thereby recovering realistic texture without compromising spatial fidelity.

\section*{Acknowledgements}

This work has been partly funded through ANR-10-IDEX-0001-02 PSL* Université Paris and from the Agence Nationale de la Recherche (ANR) under project ANR-24-CE19-3949-01 COMPURES. This work was granted access to the HPC resources of IDRIS under the allocations 2025-A0181016159, 2025-AD011016159R1, 2025-AD010617080, 2025-AD011011495R5, 2020-AD011011495 made by GENCI.

\bibliography{main}
\bibliographystyle{icml2026}

\newpage
\appendix
\onecolumn

\section{Diffusion Models Parameterization}
\label{sec:appendix_background}

\subsection{DDIM}
\label{sec:appendix_background:ddim}
We briefly recall the DDIM/DDPM (VP) forward corruption process and its common parameterization.
Let $t\in\{0,\dots,T\}$ and define a variance schedule $\{\beta_t\}_{t=1}^T$ with
$\alpha_t\triangleq 1-\beta_t$ and cumulative product $\bar{\alpha}_t\triangleq\prod_{s=1}^t\alpha_s$ (with $\bar{\alpha}_0\triangleq 1$).
The forward noising process can be written in closed form as
\begin{equation}
    \mathbf{x}_t = \sqrt{\bar{\alpha}_t}\,\mathbf{x}_0 + \sqrt{1-\bar{\alpha}_t}\,\boldsymbol{\epsilon},\qquad \boldsymbol{\epsilon}\sim\mathcal{N}(\mathbf{0},\mathbf{I}).
    \label{eq:ddim_forward_closed_form}
\end{equation}
It is often convenient to introduce the corresponding (scalar) noise level
\begin{equation}
    \sigma_t \;\triangleq\; \sqrt{\frac{1-\bar{\alpha}_t}{\bar{\alpha}_t}},
    \label{eq:ddim_sigma_def}
\end{equation}
so that the normalized variable $\tilde{\mathbf{x}}_t\triangleq \mathbf{x}_t/\sqrt{\bar{\alpha}_t}$ satisfies
$\tilde{\mathbf{x}}_t = \mathbf{x}_0 + \sigma_t\boldsymbol{\epsilon}$.

\paragraph{Noise prediction and score estimation.}
We denote by $\mathrm{Net}_\theta(\cdot, y, t)$ a base neural network that receives a (possibly rescaled) noisy input together with the conditioning $y$ and the time index $t$.
In the standard DDIM/DDPM setup, the model is trained to predict the added noise:
\begin{equation}
    \boldsymbol{\epsilon}_\theta(\mathbf{x}_t,y,t)\;\approx\;\boldsymbol{\epsilon}.
\end{equation}
A simple and commonly used input normalization is to scale the network input by
$\sqrt{\bar{\alpha}_t}=1/\sqrt{1+\sigma_t^2}$, which keeps the input scale closer to the data scale across $t$.
Concretely, we can write this as
\begin{equation}
    \boldsymbol{\epsilon}_\theta(\mathbf{x}_t,y,t)
    \;\triangleq\;
    \mathrm{Net}_\theta\!\left(\mathbf{x}_t,\,y,\,t\right).
    \label{eq:ddim_eps_net_precond}
\end{equation}
The parameters are learned with the usual denoising objective
\begin{equation}
    \theta^\star
    = \arg\min_\theta\; \mathbb{E}_{t,\,\mathbf{x}_0,\,\boldsymbol{\epsilon}}
    \left[\,\left\|\boldsymbol{\epsilon}-\boldsymbol{\epsilon}_\theta(\mathbf{x}_t,y,t)\right\|_2^2\,\right].
    \label{eq:ddim_eps_loss}
\end{equation}
This noise predictor induces an estimate of the conditional score via
\begin{equation}
    \nabla_{\mathbf{x}}\log p_t(\mathbf{x}_t\mid y)
    \;\approx\;
    -\frac{1}{\sqrt{1-\bar{\alpha}_t}}\,\boldsymbol{\epsilon}_\theta(\mathbf{x}_t,y,t),
    \label{eq:ddim_score_from_eps}
\end{equation}
(up to discretization and time-parameterization conventions).

\paragraph{Alternative parameterization (\texorpdfstring{$\mathbf{x}_0$}{x0}-prediction).}
In our experimental analysis (Sec.~\ref{sec:experiments:epsilon_vs_x0}), we also evaluate models parameterized to predict the clean signal directly, i.e., $\mathbf{x}_\theta(\mathbf{x}_t, y, t) \approx \mathbf{x}_0$.
To maintain consistency with the DDIM formulation, we derive the implicit noise estimate by inverting Eq.~\eqref{eq:ddim_forward_closed_form}:
\begin{equation}
    \boldsymbol{\epsilon}^{\mathbf{x}_0}_\theta(\mathbf{x}_t,y,t)
    \;\triangleq\;
    \frac{\mathbf{x}_t - \sqrt{\bar{\alpha}_t}\,\mathbf{x}_\theta(\mathbf{x}_t,y,t)}{\sqrt{1-\bar{\alpha}_t}}.
    \label{eq:ddim_eps_from_x0}
\end{equation}
This derived $\boldsymbol{\epsilon}^{\mathbf{x}_0}_\theta$ is then used in place of the direct prediction $\boldsymbol{\epsilon}_\theta$ for all subsequent inversion and sampling steps.

\paragraph{Deterministic DDIM step (\texorpdfstring{$\eta=0$}{eta=0}).}
Given $\boldsymbol{\epsilon}_\theta(\mathbf{x}_t,y,t)$, the corresponding estimate of the clean sample is
\begin{equation}
    \hat{\mathbf{x}}_0(\mathbf{x}_t,y,t)
    \;=\;
    \frac{\mathbf{x}_t-\sqrt{1-\bar{\alpha}_t}\,\boldsymbol{\epsilon}_\theta(\mathbf{x}_t,y,t)}{\sqrt{\bar{\alpha}_t}}.
    \label{eq:ddim_x0_hat}
\end{equation}
The deterministic DDIM update (no additional noise) is then
\begin{equation}
    \mathbf{x}_{t-1}
    = \sqrt{\bar{\alpha}_{t-1}}\,\hat{\mathbf{x}}_0(\mathbf{x}_t,y,t)
    + \sqrt{1-\bar{\alpha}_{t-1}}\,\boldsymbol{\epsilon}_\theta(\mathbf{x}_t,y,t).
    \label{eq:ddim_deterministic_step}
\end{equation}
Expanding Eq.~\eqref{eq:ddim_deterministic_step} shows that $\mathbf{x}_{t-1}$ is an affine function of $(\mathbf{x}_t,\boldsymbol{\epsilon}_\theta)$ and can therefore be written in the generic form of Eq.~\eqref{eq:general_ode_backward} with coefficients $a(t)$ and $b(t)$.

\subsection{EDM}
\label{sec:appendix_background:edm}
EDM (\cite{karras2022elucidatingdesignspacediffusionbased}) adopts a VE-style corruption process parameterized directly by a noise level $\sigma_t$.
In its simplest form, noisy inputs are generated as
\begin{equation}
    \mathbf{x}_t = \mathbf{x}_0 + \sigma_t\boldsymbol{\epsilon},\qquad \boldsymbol{\epsilon}\sim\mathcal{N}(\mathbf{0},\mathbf{I}).
    \label{eq:edm_forward}
\end{equation}

\paragraph{Noise schedule.}
For sampling (and for defining the discrete PF-ODE trajectory), EDM commonly uses the Karras noise schedule that interpolates between $(\sigma_{\max},\sigma_{\min})$ with a curvature parameter $\rho$ (typically $\rho=7$).
For $N$ steps, define $r_i\triangleq i/(N-1)$ for $i=0,\dots,N-1$, and set
\begin{equation}
    \sigma_i
    = \Big(\sigma_{\max}^{1/\rho} + r_i\big(\sigma_{\min}^{1/\rho}-\sigma_{\max}^{1/\rho}\big)\Big)^{\rho}.
    \label{eq:karras_sigma_schedule}
\end{equation}

\paragraph{Preconditioning.}
As emphasized in \cite{karras2022elucidatingdesignspacediffusionbased}, the scale of $\mathbf{x}_t$ varies substantially with $\sigma_t$, and neural networks tend to train best when inputs/targets remain on a comparable scale.
EDM therefore wraps a base network $\mathrm{Net}_\theta$ with an explicit input/output preconditioning.
In this appendix, we keep the network interface in terms of a generic time input $t$ (rather than $\sigma$); concretely, $t$ can be taken as a fixed embedding of the noise level such as
$t=c_{\mathrm{noise}}(\sigma_t)=\tfrac14\log\sigma_t$.
The preconditioned denoiser takes the form
\begin{equation}
    F_\theta(\mathbf{x}_t,y,t)
    \;\triangleq\;
    c_{\mathrm{skip}}(\sigma_t)\,\mathbf{x}_t
    + c_{\mathrm{out}}(\sigma_t)\,\mathrm{Net}_\theta\!\big(c_{\mathrm{in}}(\sigma_t)\,\mathbf{x}_t,\,y,\,t\big),
    \label{eq:edm_precond_denoiser}
\end{equation}
with coefficients
\begin{equation}
    c_{\mathrm{in}}(\sigma)=\frac{1}{\sqrt{\sigma^2+\sigma_{\mathrm{data}}^2}},\qquad
    c_{\mathrm{skip}}(\sigma)=\frac{\sigma_{\mathrm{data}}^2}{\sigma^2+\sigma_{\mathrm{data}}^2},\qquad
    c_{\mathrm{out}}(\sigma)=\frac{\sigma\,\sigma_{\mathrm{data}}}{\sqrt{\sigma^2+\sigma_{\mathrm{data}}^2}}.
    \label{eq:edm_precond_coeffs}
\end{equation}
These choices ensure that the network sees approximately \textbf{\textit{unit-variance}} inputs for all $t$ and that the overall denoiser behaves smoothly across noise levels.

\paragraph{Training objective and score connection.}
EDM trains the model to predict the clean sample (denoised data) with a weighted least-squares loss:
\begin{equation}
    \theta^\star
    = \arg\min_\theta\;\mathbb{E}_{t,\,\mathbf{x}_0,\,\boldsymbol{\epsilon}}
    \left[\,w(\sigma_t)\,\left\|F_\theta(\mathbf{x}_t,y,t)-\mathbf{x}_0\right\|_2^2\,\right],
    \label{eq:edm_denoiser_loss}
\end{equation}
where a standard choice is
\begin{equation}
    w(\sigma)=\frac{\sigma^2+\sigma_{\mathrm{data}}^2}{(\sigma\,\sigma_{\mathrm{data}})^2}.
    \label{eq:edm_loss_weight}
\end{equation}
With a denoiser parameterization, the score can be recovered through the standard relation
\begin{equation}
    \nabla_{\mathbf{x}}\log p_t(\mathbf{x}_t\mid y)
    \;\approx\;
    \frac{F_\theta(\mathbf{x}_t,y,t)-\mathbf{x}_t}{\sigma_t^2},
    \label{eq:edm_score_from_denoiser}
\end{equation}
again up to discretization conventions.
\section{Metrics}
\label{sec:appendix_metrics}

This appendix provides a complete description of the three quantitative scores used throughout the paper to assess inversion quality and the trade-off between realism and structural fidelity.
Given an experiment producing (i) an inverted latent $\mathbf{x}_T$ and (ii) a generated image $\hat{\mathbf{x}}$ conditioned on an input $\mathbf{c}$ (e.g., sketch, masked image, or low-resolution observation), we report:
(i) a \emph{Gaussianity} score $S_{\mathrm{G}}$ for $\mathbf{x}_T$,
(ii) a \emph{high-frequency realism} score $S_{\mathrm{HF}}$ for $\hat{\mathbf{x}}$, and
(iii) a \emph{low-frequency structural fidelity} score $S_{\mathrm{LF}}$ between $\hat{\mathbf{x}}$ and $\mathbf{c}$.
All scores are standardized to lie in $[0,1]$ with higher values indicating better behavior.

\subsection{Score definitions and dataset-specific settings}
\label{subsec:appendix_scores}

We use three scores with different aggregation levels.
First, $S_{\mathrm{G}}(\mathcal{D})$ is a \emph{dataset-level} Gaussianity score computed from the collection of terminal latents produced by inversion on a dataset $\mathcal{D}$.
Second, $S_{\mathrm{HF}}(\hat{\mathbf{x}}\,;\mathcal{D})$ is a \emph{per-sample} high-frequency realism score that compares a generated sample $\hat{\mathbf{x}}$ to a dataset-wide reference statistic computed on real images from $\mathcal{D}$.
Third, $S_{\mathrm{LF}}(\hat{\mathbf{x}},\mathbf{c})$ is a \emph{per-sample} structural fidelity score computed between a generated output $\hat{\mathbf{x}}$ and its corresponding conditioning input $\mathbf{c}$.
Dataset-specific choices (e.g., color handling, masking, and frequency cutoffs) are introduced locally in the relevant metric sections below.

\subsection{Gaussianity of the inverted latent $S_{\mathrm{G}}$}
\label{subsec:sg}

\begin{figure}[t]
    \centering
    \subfloat[VAE Latent]{
        \includegraphics[width=0.45\columnwidth]{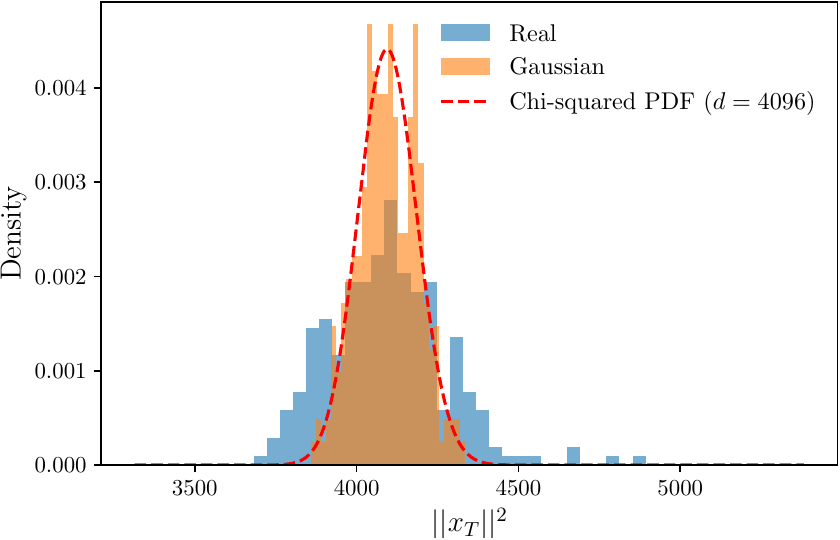}
        \label{fig:appendix_sg_vae}
    }\hfill
    \subfloat[Pixel Space]{
        \includegraphics[width=0.45\columnwidth]{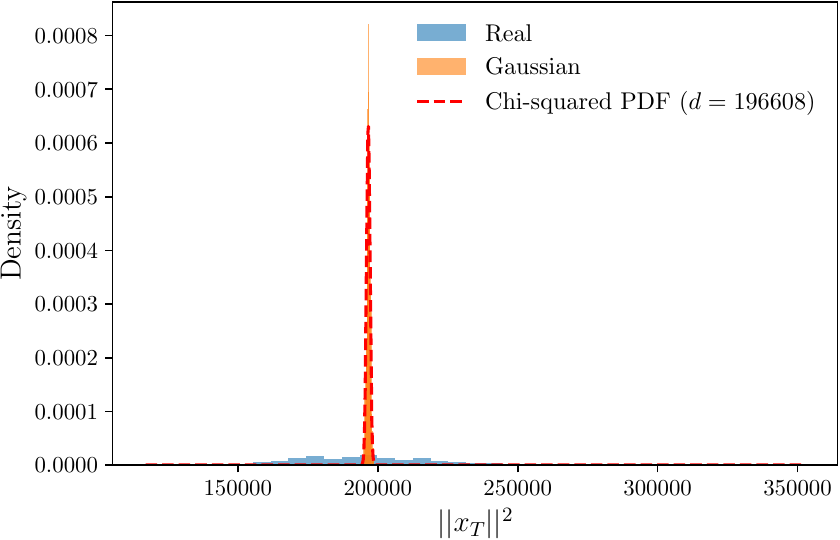}
        \label{fig:appendix_sg_pixel}
    }
    \caption{Empirical distributions of squared norms $\|\mathbf{x}_T\|_2^2$ (histograms) compared to the theoretical $\chi^2(d)$ distribution (orange curve) for (a) VAE latents ($d=4096$) and (b) pixel space ($d=196608$) on BBBC021 for standard DDIM inversion.}
    \label{fig:appendix_sg_chi2_examples}
\end{figure}

\paragraph{Definition via a one-sample KS test.}
A core modeling assumption is that the terminal latent follows an isotropic Gaussian,
$\mathbf{x}_T\sim\mathcal{N}(\mathbf{0}_d,\mathbf{I}_d)$,
where $d$ is the latent dimensionality (flattening channels and spatial positions).
For a sample $\mathbf{x}_T$, define the squared norm
\begin{equation}
    r \;\triangleq\; \|\mathbf{x}_T\|_2^2.
\end{equation}
If $\mathbf{x}_T\sim\mathcal{N}(\mathbf{0}_d,\mathbf{I}_d)$, then $r\sim\chi^2(d)$.
Given a collection of inverted latents $\{\mathbf{x}_T^{(i)}\}_{i=1}^N$, we compute $\{r_i\}_{i=1}^N$ and perform a one-sample Kolmogorov--Smirnov (KS) test between the empirical CDF of $\{r_i\}$ and the CDF of $\chi^2(d)$.
Let $F_N$ be the empirical CDF of $\{r_i\}_{i=1}^N$ and let $F_{\chi^2(d)}$ be the CDF of the $\chi^2(d)$ distribution.
The one-sample KS statistic is
\begin{equation}
    D_{\mathrm{KS}} \;\triangleq\; \sup_{u\in\mathbb{R}} \big|F_N(u) - F_{\chi^2(d)}(u)\big| \in [0,1].
\end{equation}
We define the Gaussianity score as
\begin{equation}
    S_{\mathrm{G}} \;\triangleq\; 1 - D_{\mathrm{KS}}.
    \label{eq:sg_def}
\end{equation}
Thus, $S_{\mathrm{G}}\approx 1$ indicates that the \emph{radial distribution} of $\mathbf{x}_T$ matches the Gaussian prior, while lower values indicate non-Gaussianity.
We also record the KS $p$-value as a diagnostic.
In the appendix we additionally report illustrative examples in Figure \ref{fig:appendix_sg_chi2_examples} for both VAE latent space and pixel space on BBBC021 for standard DDIM inversion.

\paragraph{Why an additional dependence probe is useful.}
The KS test in Eq.~\eqref{eq:sg_def} only constrains the \emph{marginal distribution} of $\|\mathbf{x}_T\|_2^2$.
It can miss strong \emph{local dependencies} (e.g., spatially correlated patches) that may coexist with an approximately correct radial distribution.
To complement $S_{\mathrm{G}}$, we therefore introduce $S_{DC}$, a decorrelation metric that measures the fraction of unusually smooth patches in $\mathbf{x}_T$.

\subsection{Decorrelation patch ratio}
\label{subsec:sg_patch}

For a latent tensor $\mathbf{x}_T\in\mathbb{R}^{C\times H\times W}$, we slide a $k\times k$ window over spatial locations and compute a simple \emph{patch smoothness} statistic, in other words, a local autocorrelation measure.
Let $\mathcal{P}_{u}$ denote the set of indices in the patch centered at location $u$.
We define the (channel-wise) patch roughness as the average squared finite difference inside the patch,
\begin{equation}
    \rho_c(u) \;\triangleq\; \frac{1}{|\mathcal{E}_u|}\sum_{(p,q)\in\mathcal{E}_u}\big(z_{c,p}-z_{c,q}\big)^2,
\end{equation}
where $\mathcal{E}_u$ collects horizontal/vertical neighboring pairs inside the patch.
Intuitively, i.i.d. Gaussian latents produce patches with a predictable roughness distribution, whereas spatially correlated latents yield unusually \emph{smooth} patches.

Because inversion may be performed either directly in pixel space or in a learned autoencoder latent space, the dimensionality $(C,H,W)$ can vary across models and datasets.
We therefore calibrate the null distribution (and the threshold $\tau$ below) for each admissible latent shape encountered in our experiments, ensuring comparable false-positive rates across settings.

We calibrate a threshold $\tau$ by Monte Carlo under the null hypothesis $\mathbf{z}\sim\mathcal{N}(\mathbf{0},\mathbf{I})$ so that
\begin{equation}
    \mathbb{P}(\rho_c(u) \le \tau) \approx \mathrm{FPR}
\end{equation}
for a target false-positive rate (we use $\mathrm{FPR}=0.05$).
We then flag a patch as \emph{autocorrelated} if a sufficient fraction of channels simultaneously satisfy $\rho_c(u)\le\tau$.
Finally, we report the \emph{autocorrelated patch ratio}
\begin{equation}
    S_{DC} \;\triangleq\; \frac{1}{|\Omega|}\sum_{u\in\Omega}\mathbf{1}\{\text{patch at }u\text{ is flagged}\}\in[0,1],
\end{equation}
where $\Omega$ is the set of valid patch centers.
Large $S_{DC}$ indicates substantial local dependencies (patch-wise structure) in $\mathbf{x}_T$.
In the appendix we provide (i) qualitative examples illustrating how patches are flagged, and (ii) scatter plots relating $S_{DC}$ to $S_{\mathrm{G}}(\mathcal{D})$.
\begin{figure}[t]
    \centering
    \subfloat[VAE Latent]{
        \includegraphics[width=0.45\columnwidth]{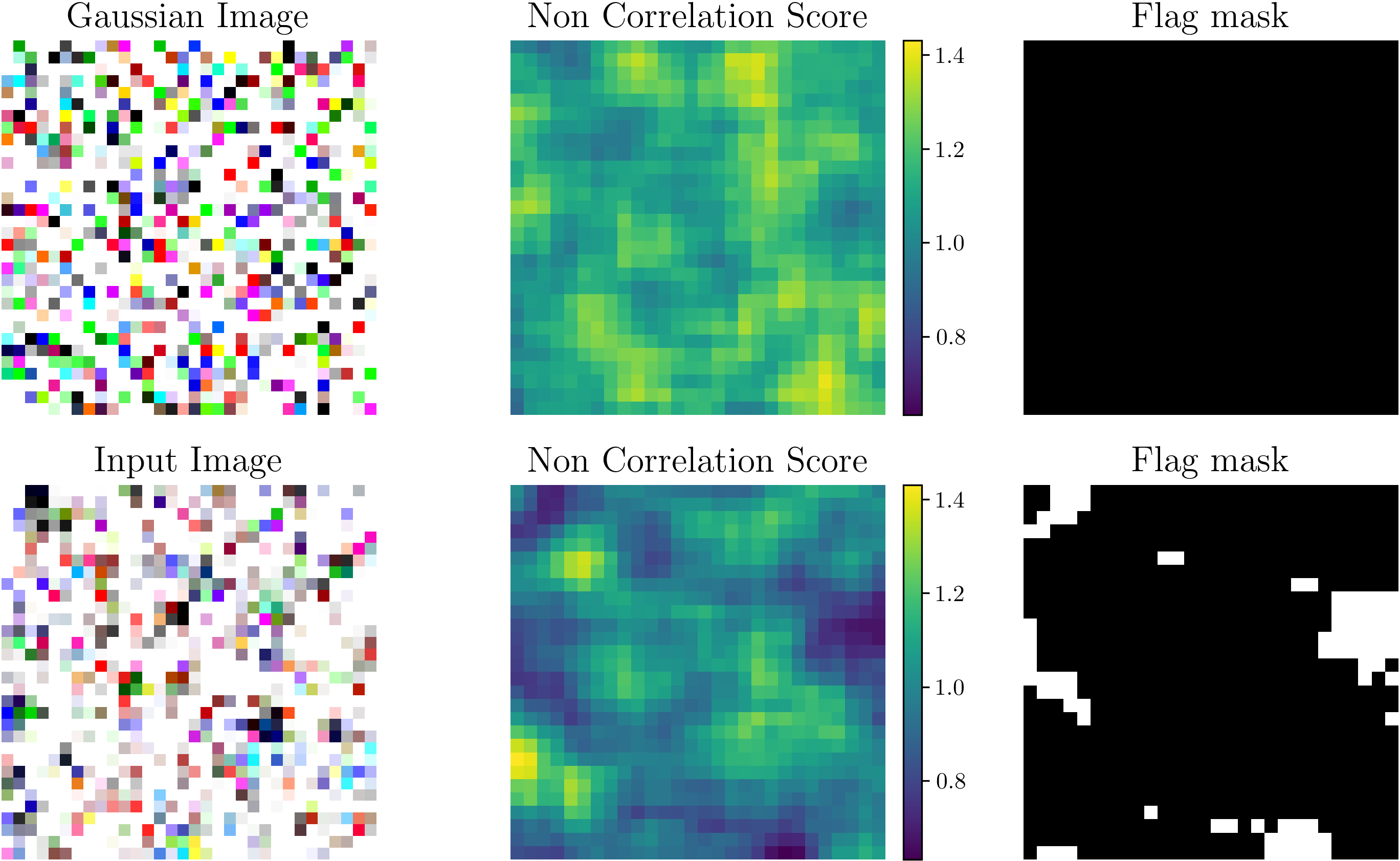}
        \label{fig:appendix_sg_patch_vae}
    }\hfill
    \subfloat[Pixel Space]{
        \includegraphics[width=0.45\columnwidth]{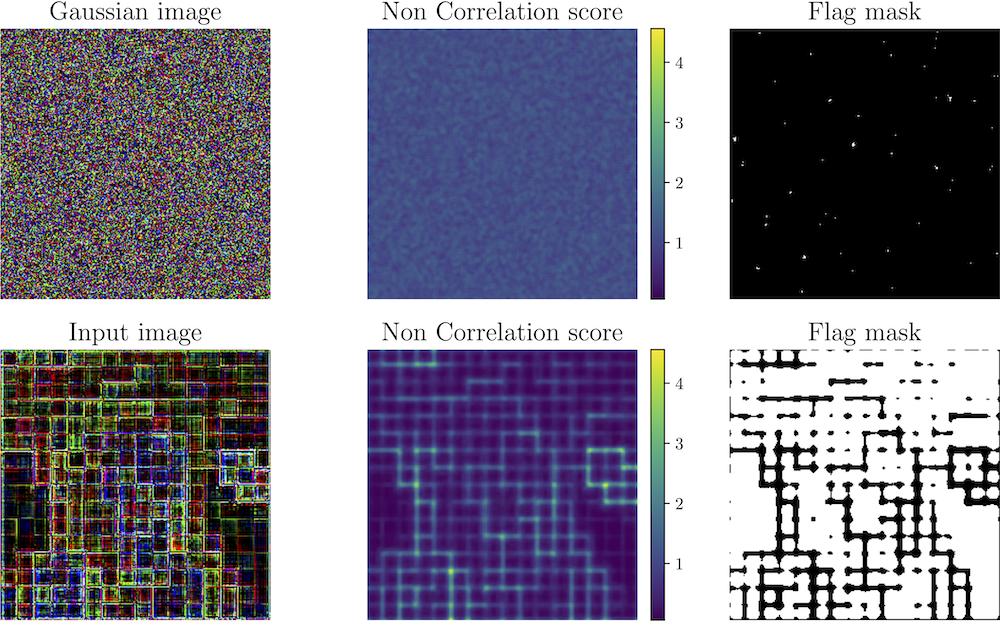}
        \label{fig:appendix_sg_patch_pixel}
    }
    \caption{Scatter plots of the decorrelation patch ratio $S_{DC}$ versus the Gaussianity score $S_{\mathrm{G}}$ for (a) VAE latents ($d=4096$) and (b) pixel space ($d=196608$) on BBBC021 for standard DDIM inversion. Each point corresponds to a different inversion run with varying seeds and number of steps. The score $S_{DC}$ is the ratio of white patchs in the Flag Mask image.}
    \label{fig:appendix_sg_patch_scatter}
\end{figure}

\begin{figure}[t]
    \centering
    \includegraphics[width=0.9\linewidth]{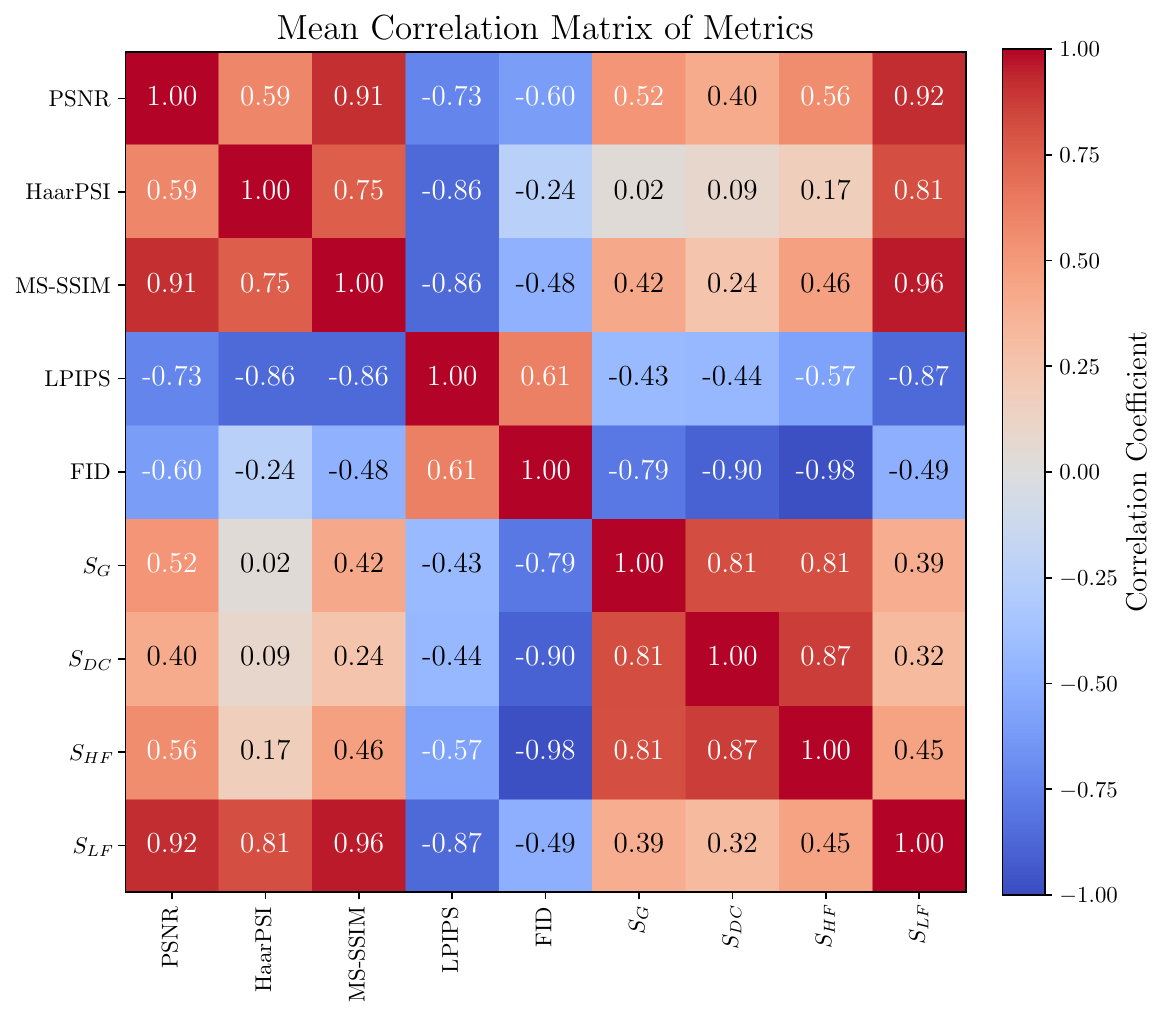}
    \caption{Correlation metrics between all metrics for different methods on BBBC021 and Edges2Shoes datasets.}
    \label{fig:metric_correlation_matrix}
\end{figure}

\begin{figure}[H]
    \centering
    \includegraphics[width=0.45\columnwidth]{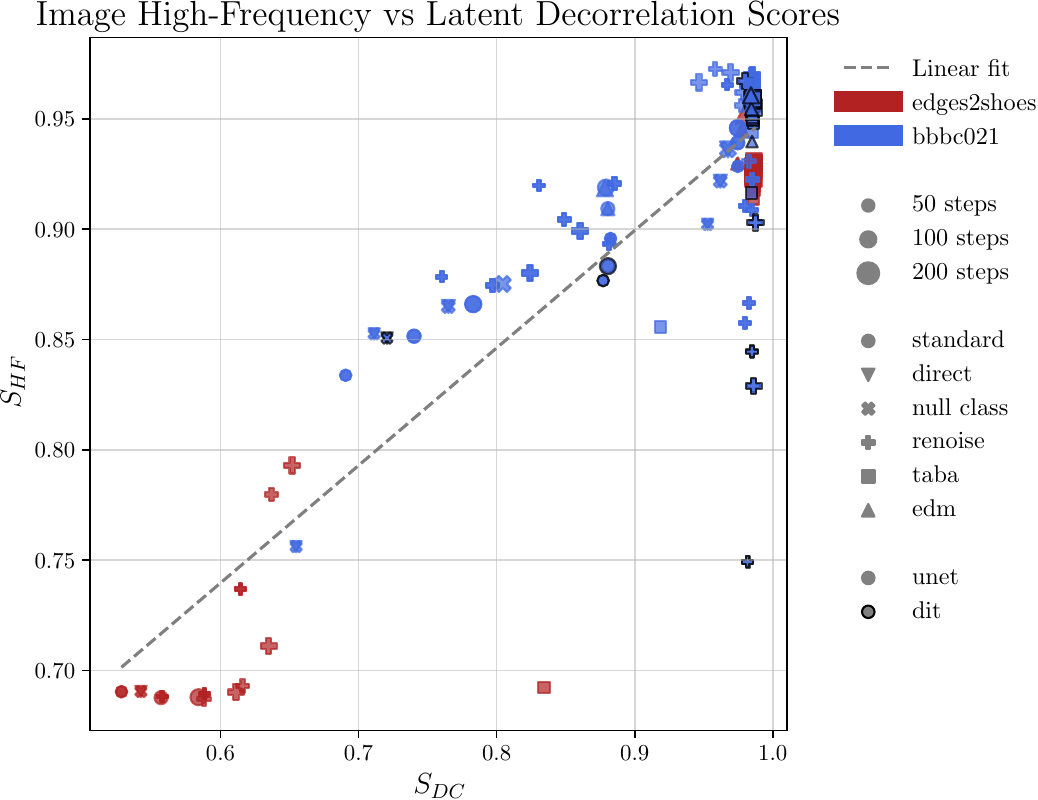}
    \caption{Variation of the High-Frequency Score $S_{\mathrm{HF}}$ against the Latent Decorrelation Score $S_{\mathrm{DC}}$ across different methods and datasets. A strong positive correlation is observed, indicating that as the patch correlation in the latent increases (higher spectral collapse), the high-frequency content in the generated images decreases.}
    \label{app:fig:shf_vs_sdc}
\end{figure}

\paragraph{Correlation with $S_{\mathrm{G}}$.} Across our experiments, we observe a clear positive correlation between $S_{DC}$ and $S_{\mathrm{G}}$ (see Figure~\ref{fig:metric_correlation_matrix}). In Figure~\ref{app:fig:shf_vs_sdc} we provide scatter plots of $S_{DC}$ against $S_{\mathrm{G}}$ for BBBC021 in both VAE latent space for various methods and architectures.

\subsection{High-frequency realism score $S_{\mathrm{HF}}$}
\label{subsec:shf}

Spectral collapse manifests as a deficit of high-frequency energy in generated images.
We quantify this deficit by comparing the high-frequency wavelet energy of a generated image to the typical high-frequency energy of real images from the same dataset.

\begin{figure}[H]
    \centering
    \includegraphics[width=0.9\linewidth]{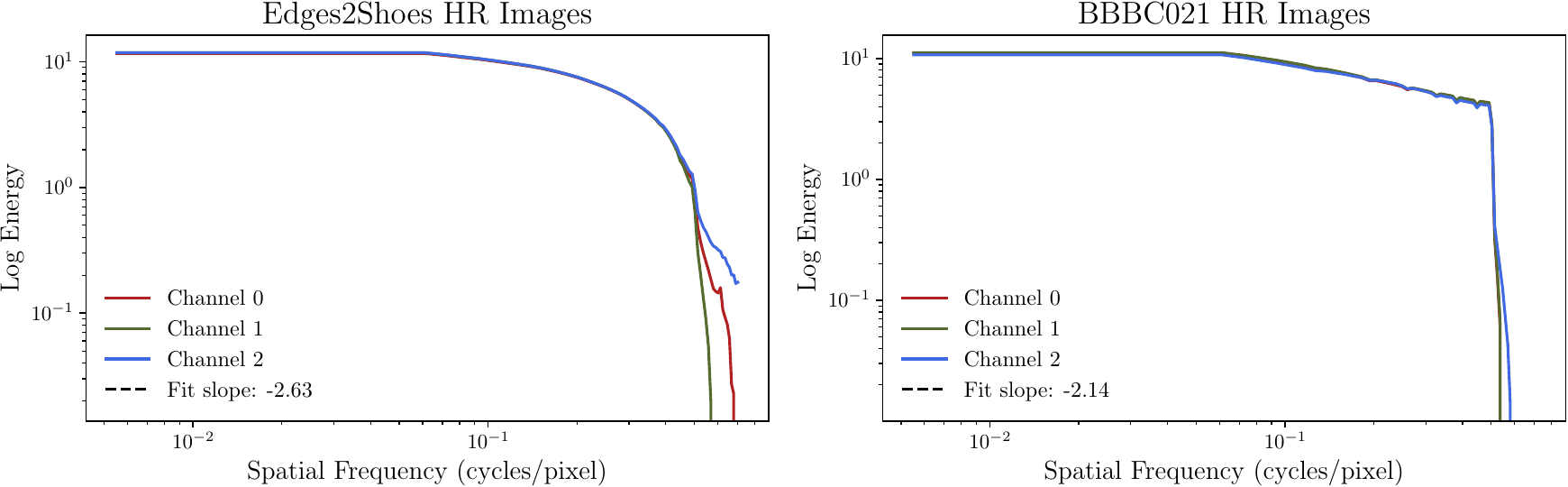}
    \caption{Radially averaged log power spectra of real images from our datasets, with linear fit to estimate the moment of the wavelet.}
    \label{fig:appendix_fft_profiles}
\end{figure}

\paragraph{Step 1: choosing a wavelet family from the dataset spectrum.}
We estimate the average log power spectrum of real images by computing FFTs, taking the radially averaged log-energy as a function of spatial frequency (cycles per pixel), and fitting a line over a mid-frequency band (we use $[0.005,0.4]$ cycles/pixel).
Let $s$ be the fitted slope (averaged over channels). We then use the Symmlet family \texttt{sym}$k$ with
\begin{equation}
    k \;\triangleq\; \left\lceil |s| \right\rceil.
\end{equation}
Across our datasets the fitted slope is typically close to $-3.5$, yielding \texttt{sym4}.
We provide the corresponding radial FFT profiles and fitted slopes for each dataset in Figure~\ref{fig:appendix_fft_profiles}.

\paragraph{Step 2: reference log-energy on real images.}
Fix a wavelet (typically \texttt{sym4}) and a number of levels $L$.
For an image $\mathbf{x}$, let $\mathcal{W}_\ell(\mathbf{x})$ denote the collection of detail coefficients at level $\ell$ (high-pass bands).
We define the finest-scale high-frequency energy
\begin{equation}
    E_{\mathrm{HF}}(\mathbf{x}) \;\triangleq\; \sum_{c=1}^C \|\mathcal{W}_{1}(\mathbf{x}_c)\|_2^2,
\end{equation}
and its log-energy $\log E_{\mathrm{HF}}(\mathbf{x})$.
Given a dataset of real images $\mathcal{D}=\{\mathbf{x}_i\}_{i=1}^N$, we compute the reference
\begin{equation}
    \mu_{\mathrm{HF}}(\mathcal{D}) \;\triangleq\; \frac{1}{N}\sum_{i=1}^N \log E_{\mathrm{HF}}(\mathbf{x}_i).
    \label{eq:hf_ref_mu}
\end{equation}

\begin{figure}[H]
    \centering
    \includegraphics[width=0.9\linewidth]{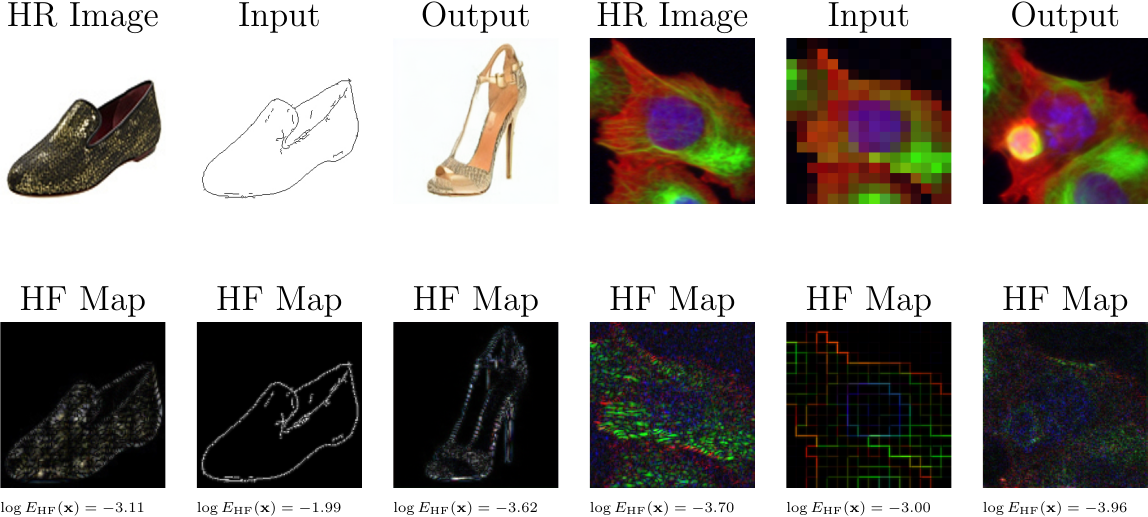}
    \caption{Wavelet detail maps at the finest scale (level 1) for representative images from BBBC021 Edges2Shoes and CelebAMask-HQ, illustrating the types of high-frequency content captured with the average log-energy $E_{\mathrm{HF}}$.}
    \label{fig:walevet_details_hf}
\end{figure}

We additionally visualize representative wavelet detail maps and the resulting reference energy profiles in Figure~\ref{fig:walevet_details_hf}.

\paragraph{Step 3: scoring a generated image.}
For a generated image $\hat{\mathbf{x}}$, we compute $\log E_{\mathrm{HF}}(\hat{\mathbf{x}})$ and compare it to $\mu_{\mathrm{HF}}(\mathcal{D})$.
We define
\begin{equation}
    S_{\mathrm{HF}}(\hat{\mathbf{x}})
    \;\triangleq\;
    1 - \mathrm{clip}\!\left(
    \frac{\big|\log E_{\mathrm{HF}}(\hat{\mathbf{x}})-\mu_{\mathrm{HF}}(\mathcal{D})\big|}{\big|\mu_{\mathrm{HF}}(\mathcal{D})\big|},\,0,\,1
    \right),
    \label{eq:shf_def}
\end{equation}
where $\mathrm{clip}(\cdot,0,1)$ truncates values outside $[0,1]$.
Thus, $S_{\mathrm{HF}}\approx 1$ indicates typical high-frequency energy, whereas $S_{\mathrm{HF}}\approx 0$ indicates a strong high-frequency deficit or excess.

\begin{figure}[H]
    \centering
    \includegraphics[width=0.9\linewidth]{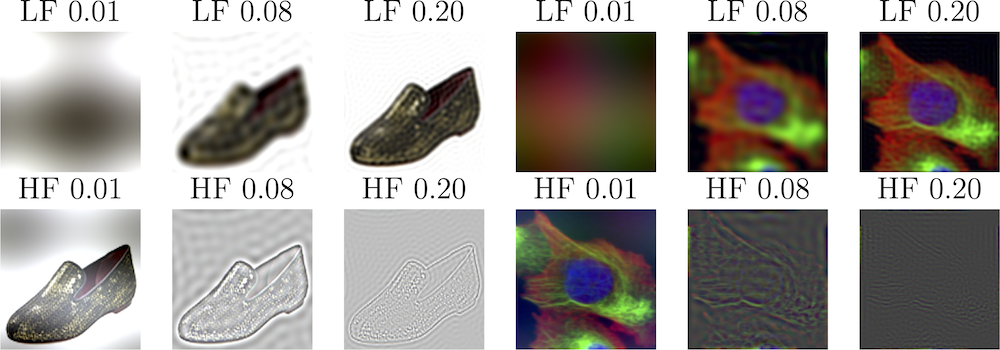}
    \caption{Examples of Fourier-filtered samples from real images in BBBC021 and Edges2Shoes. Top row: low-frequency filtered images. Bottom row: high-frequency filtered images.}
    \label{fig:fourier_filtered_samples}
\end{figure}

\begin{figure}[H]
    \centering
    \includegraphics[width=0.9\linewidth]{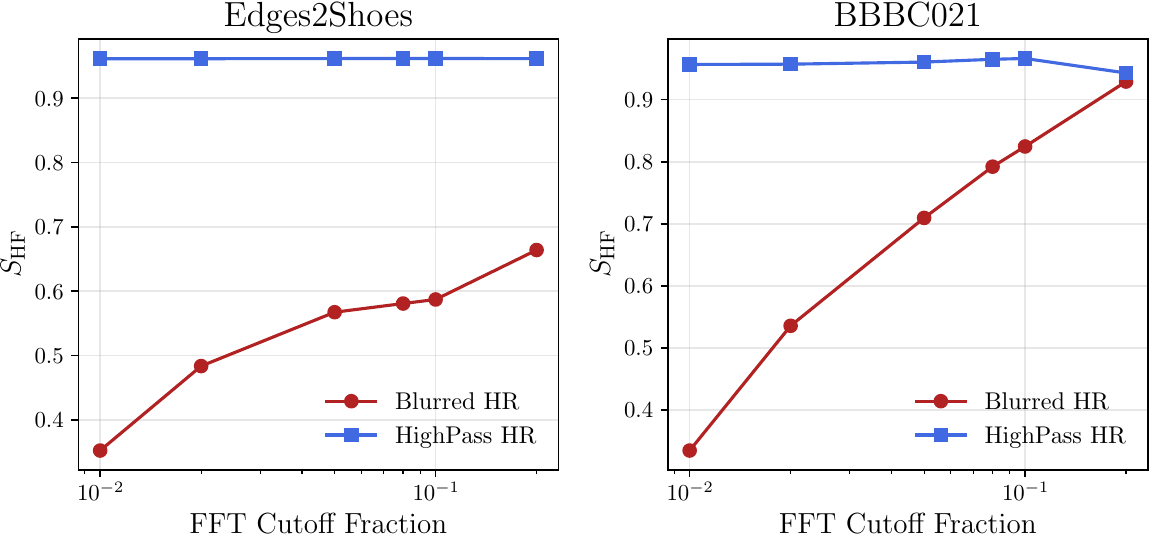}
    \caption{High-frequency realism score $S_{\mathrm{HF}}$ computed on Fourier-filtered samples from BBBC021 and Edges2Shoes. $S_{\mathrm{HF}}$ remains high and stable across low-frequency filtered images, while it significantly drops for high-frequency filtered images, confirming its sensitivity to high-frequency content.}
    \label{fig:fourier_filtered_metric}
\end{figure}

We show qualitatively how $S_{\mathrm{HF}}$ captures spectral collapse in Figure~\ref{fig:fourier_filtered_metric} and is not sensitive to how the structure of the image changes. In Figure~\ref{fig:fourier_filtered_samples} we illustrate different sets of samples on which we compute $S_{\mathrm{HF}}$. For each dataset, we compute a high-frequency filter and low-frequency filter in the Fourier domain, and show samples obtained by applying these filters to real images in Figure~\ref{fig:fourier_filtered_samples}. We can see in Figure~\ref{fig:fourier_filtered_metric} that $S_{\mathrm{HF}}$ is constant across all images obtained by low-frequency filtering, while it significantly drops for images obtained by high-frequency filtering, confirming that $S_{\mathrm{HF}}$ is indeed sensitive to high-frequency content but not to low-frequency structure.


\subsection{Low-frequency structural fidelity score $S_{\mathrm{LF}}$}
\label{subsec:slf}

\begin{figure}[t]
    \centering
    \subfloat[Edges2Shoes Low Difference]{
        \includegraphics[width=0.48\columnwidth]{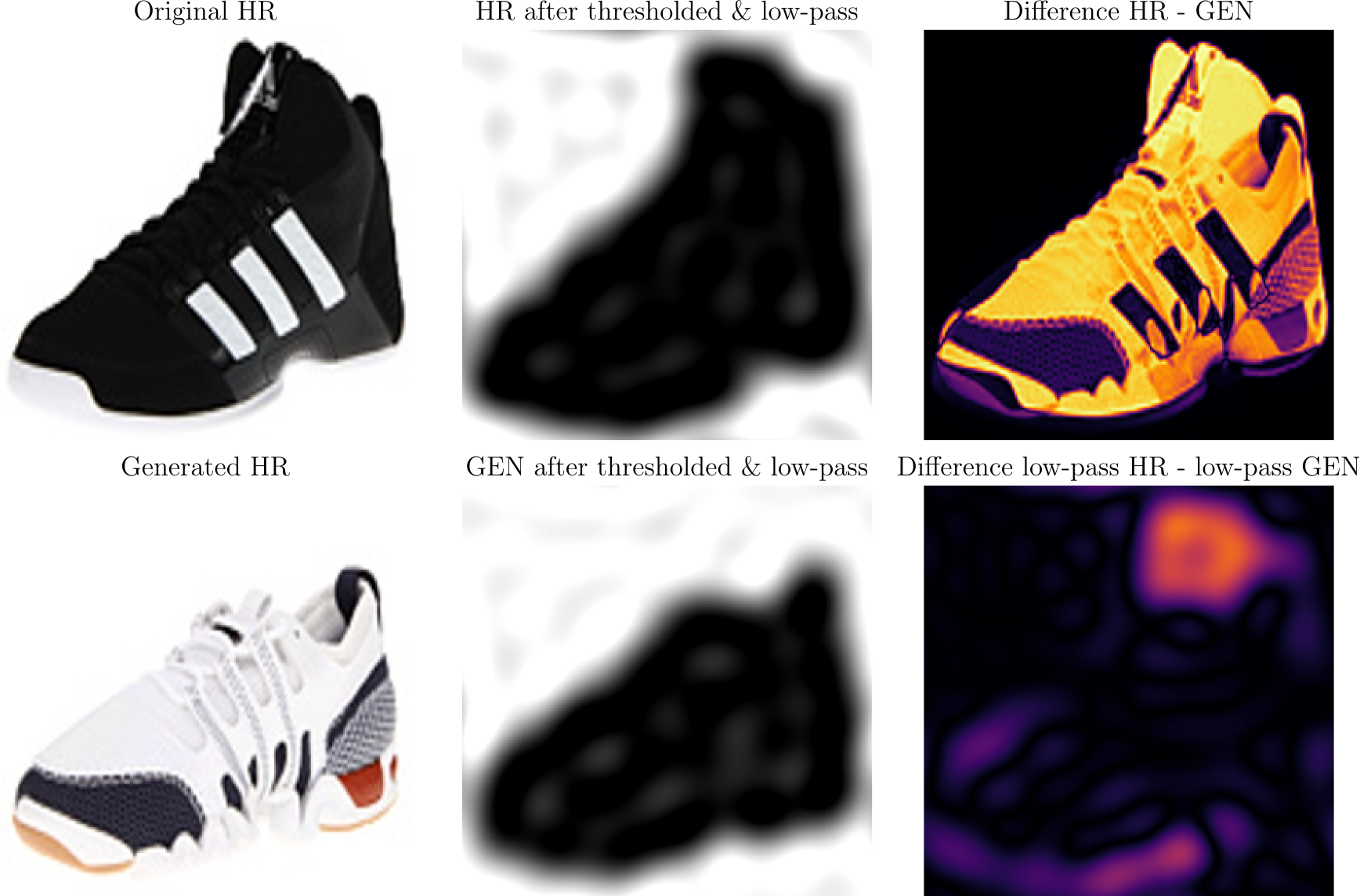}
        \label{fig:appendix_slf_edges2shoes_low}
    }\hfill
    \subfloat[Edges2Shoes High Difference]{
        \includegraphics[width=0.48\columnwidth]{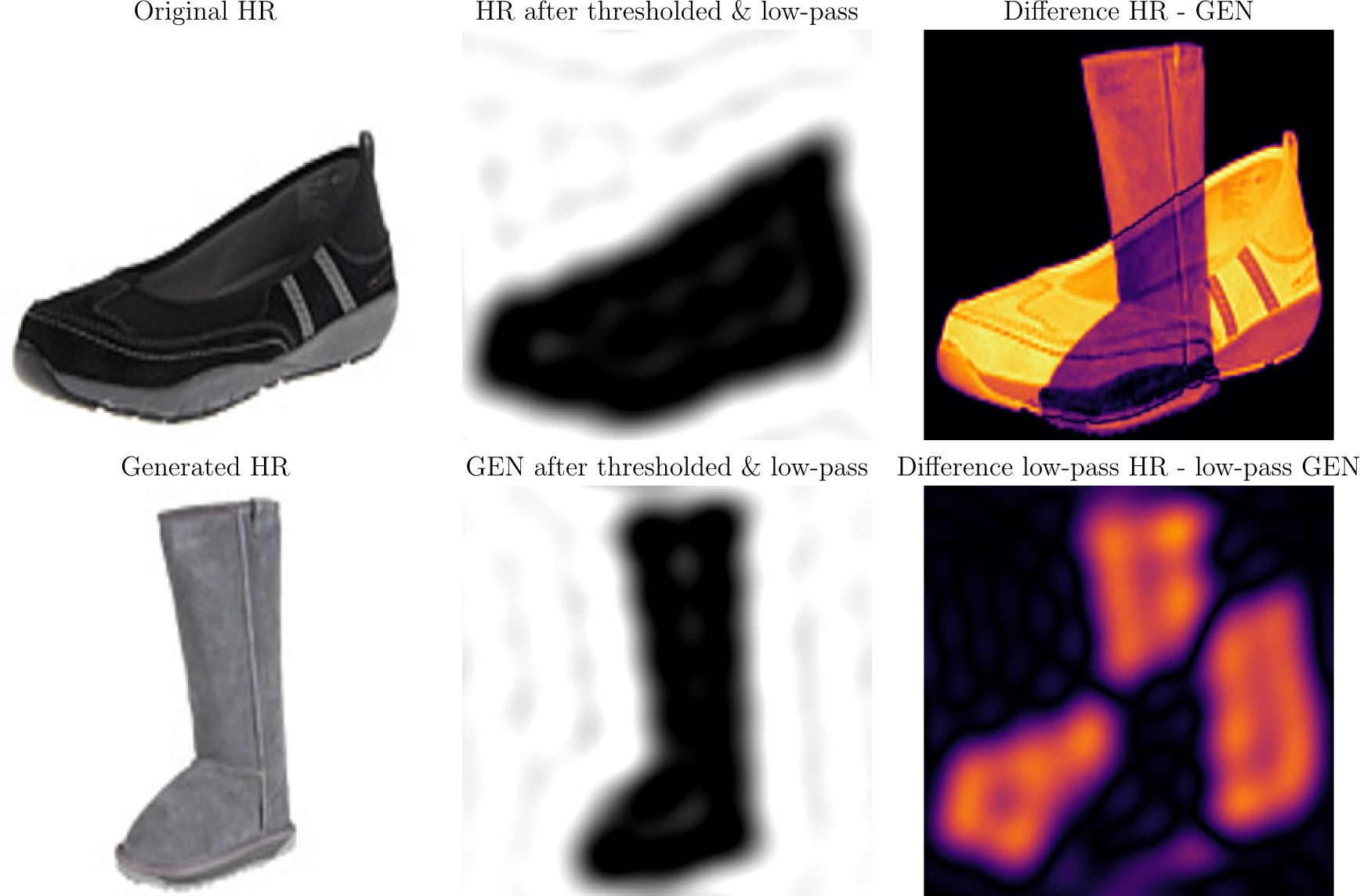}
        \label{fig:appendix_slf_edges2shoes_high}
    }

    \subfloat[BBBC021 Low Difference]{
        \includegraphics[width=0.48\columnwidth]{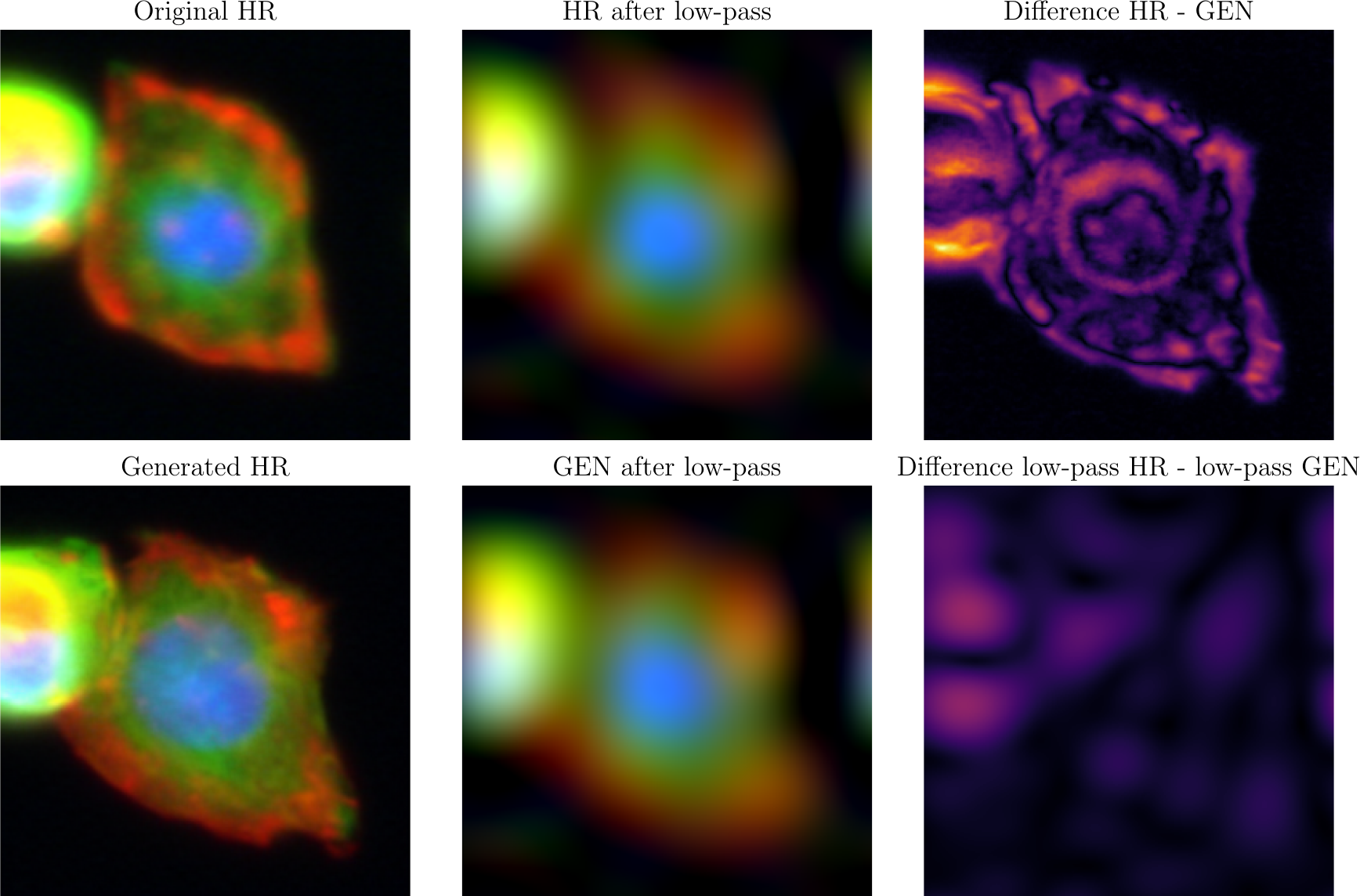}
        \label{fig:appendix_slf_bbbc021_low}
    }
    \hfill
    \subfloat[BBBC021 High Difference]{
        \includegraphics[width=0.48\columnwidth]{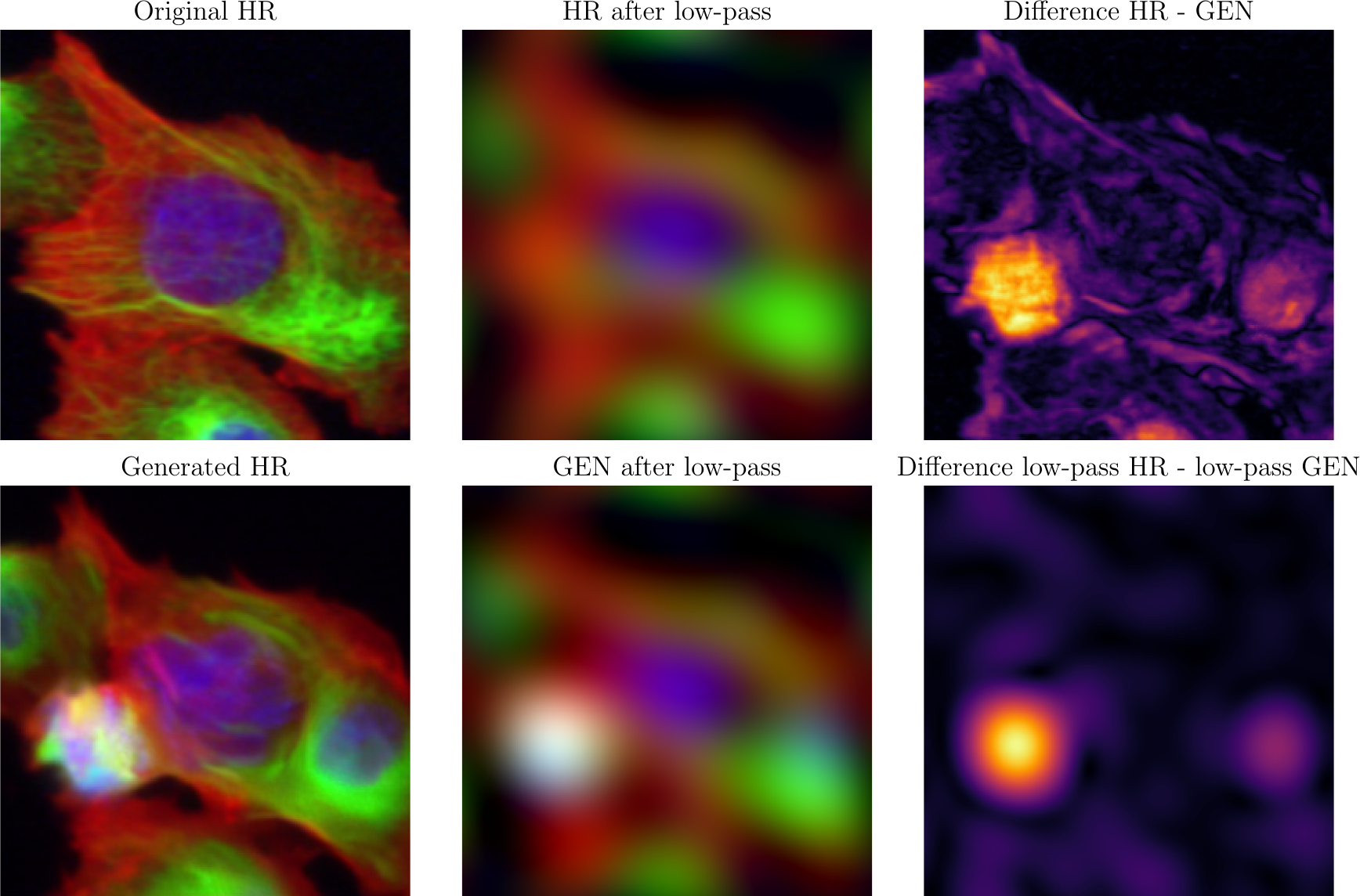}
        \label{fig:appendix_slf_bbbc021_high}
    }
    \caption{Illustration of the pipeline to compute the low-frequency structural fidelity score $S_{\mathrm{LF}}$ on (a) Edges2Shoes and (b) BBBC021. For each dataset, we show the expected output $\mathbf{c}$, the generated image $\hat{\mathbf{x}}$, their low-pass filtered versions. For Edges2Shoes, we also show apply a grayscale conversion and a mask to focus on the shape defined by the image. The SSIM score is computed between the two low-passed images. We show the MSE between the two low-passed images to illustrate the difference in low-frequency content and between the two examples to show how $S_{\mathrm{LF}}$ captures structural fidelity. The pipeline is shown for two examples per dataset: one with high $S_{\mathrm{LF}}$ (low difference) and one with low $S_{\mathrm{LF}}$ (high difference).}
    \label{fig:appendix_slf_examples}
\end{figure}

High-frequency realism should not come at the expense of global structure.
We measure structural fidelity by comparing \emph{low-frequency} content between the generated image $\hat{\mathbf{x}}$ and the input condition $\mathbf{c}$. First, depending on the dataset, we might preprocess the images. In particular for Edges2Shoes, we convert both images to grayscale before further processing. Then we mask on the non-white pixels so that the metric focuses on the shape defined by the edges. Next, we apply ideal low-pass filtering in the Fourier domain to both images, and finally compute the Structural Similarity Index (SSIM) between the resulting low-passed images. We show an illustration of the full pipeline in Figure~\ref{fig:appendix_slf_examples} for both Edges2Shoes and BBBC021 datasets.

\paragraph{Low-pass filtering in the Fourier domain.}
Let $\mathcal{F}$ denote the 2D FFT per channel.
Given an image $\mathbf{x}$, we compute $\mathcal{F}(\mathbf{x})$ and apply an ideal circular low-pass mask
\begin{equation}
    M(\omega) \;\triangleq\; \mathbf{1}\{\|\omega\|_2 \le \kappa\,\omega_{\max}\},
\end{equation}
where $\omega_{\max}$ is the Nyquist frequency and $\kappa\in(0,1)$ is the cutoff fraction (we use $\kappa=0.06$).
We denote by $\hat{\mathbf{x}}^{\mathrm{LP}}$ the low-passed generated image and by $\mathbf{c}^{\mathrm{LP}}$ the low-passed conditioning input.
\begin{equation}
    \mathbf{x}^{\mathrm{LP}} \;\triangleq\; \mathcal{F}^{-1}\big(M\odot \mathcal{F}(\mathbf{x})\big).
\end{equation}

\paragraph{SSIM on low-passed images.}
We then compute the Structural Similarity Index (SSIM) between $\hat{\mathbf{x}}^{\mathrm{LP}}$ and $\mathbf{c}^{\mathrm{LP}}$.
For two images $u$ and $v$, SSIM is defined as
\begin{equation}
    \mathrm{SSIM}(u,v)
    = \frac{(2\mu_u\mu_v + C_1)(2\sigma_{uv}+C_2)}{(\mu_u^2+\mu_v^2+C_1)(\sigma_u^2+\sigma_v^2+C_2)},
\end{equation}
where $\mu$ and $\sigma^2$ denote local means and variances, $\sigma_{uv}$ the local covariance, and $C_1,C_2$ are stabilizing constants.
In typical image settings SSIM lies in $[0,1]$ and is commonly interpreted as a perceptual similarity measure sensitive to luminance/contrast/structure.

We define the low-frequency score as
\begin{equation}
    S_{\mathrm{LF}} \;\triangleq\; \mathrm{SSIM}\big(\hat{\mathbf{x}}^{\mathrm{LP}},\,\mathbf{c}^{\mathrm{LP}}\big),
    \label{eq:slf_def}
\end{equation}
optionally averaged over channels.
Because we work on low-passed images, $S_{\mathrm{LF}}$ emphasizes global structure (silhouette, large shapes, spatial alignment) and de-emphasizes texture.

\paragraph{Dataset semantics (why color may or may not matter).}
On BBBC021, color encodes biological staining and is semantically meaningful; we compute SSIM in RGB. On Edges2Shoes, the condition is an edge map and does not specify a unique color; we compute SSIM on grayscale and thresholed low-passed images to focus on shape as illustrated in Figure~\ref{fig:appendix_slf_examples}.

\section{Theoretical Analysis: Spectral Stability of Inversion}
\label{sec:appendix_theory}

In this section, we formally derive the conditions under which deterministic inversion suffers from \textit{spectral collapse}. We prove that for spectrally sparse inputs, the Probability Flow ODE (PF-ODE) vector field vanishes in high-frequency dimensions under $\boldsymbol{\epsilon}$-parameterization due to the spectral bias of deep ReLU networks. Conversely, we show that $\mathbf{x}_0$-parameterization structurally preserves variance via residual connections.

\subsection{Signal Model and Inversion Dynamics}
\label{sec:appendix_theory:setup}

Let $\{\mathbf{x}_t\}_{t=0}^T$ denote the trajectory of the deterministic PF-ODE from data ($t=0$) to noise ($t=T$). We analyze the stability of the inversion mapping $\Phi: \mathbf{x}_0 \mapsto \mathbf{x}_T$ for a \textbf{spectrally sparse input} $\mathbf{x}_{\text{in}}$. We model $\mathbf{x}_{\text{in}}$ as a superposition of a dominant low-frequency signal $\boldsymbol{\mu}_L$ and a microscopic high-frequency perturbation $\boldsymbol{\delta}_H$:
\begin{equation}
    \mathbf{x}_{\text{in}} = \boldsymbol{\mu}_L + \boldsymbol{\delta}_H, \quad \text{with } \|\boldsymbol{\delta}_H\|_2 \ll \|\boldsymbol{\mu}_L\|_2.
\end{equation}
Here, $\boldsymbol{\mu}_L$ represents the semantic content (e.g., flat regions) and $\boldsymbol{\delta}_H$ represents the specific latent noise realization $\boldsymbol{\epsilon}$ that must be recovered. Successful inversion requires the ODE to amplify this microscopic variance such that $\mathbf{x}_T \sim \mathcal{N}(\mathbf{0}, \mathbf{I})$.

The dynamics of the Variance Preserving (VP) PF-ODE are governed by the score estimator $\nabla_{\mathbf{x}} \log p_t(\mathbf{x}_t)$. In continuous time (following the notation in Sec.~\ref{sec:appendix_background:ddim}):
\begin{equation}
    \frac{d\mathbf{x}_t}{dt} = -\frac{1}{2}\beta(t) \left[ \mathbf{x}_t + 2 \nabla_{\mathbf{x}} \log p_t(\mathbf{x}_t) \right],
    \label{eq:pf_ode_general}
\end{equation}
where $\beta(t)$ is the continuous-time variance schedule.

\subsection{Universal Spectral Bias in ReLU Networks}
Our derivation relies on the spectral decay properties of Neural Networks established by \citet{rahaman2019spectralbiasneuralnetworks}.

\textbf{Lemma 1 (Spectral Decay and Convergence Rate).}
\textit{Let $f_\theta: \mathbb{R}^d \to \mathbb{R}$ be a deep ReLU network trained via gradient descent. Its Fourier spectrum $\tilde{f}_\theta(k)$ decays polynomially with frequency magnitude $k = \|\mathbf{k}\|$. As derived in Theorem 1 (Eq. 6) of \citet{rahaman2019spectralbiasneuralnetworks}:}
\begin{equation}
    \tilde{f}_{\theta}(k) = \sum_{n=0}^{d} \frac{C_n(\theta, k) 1_{H_n}(k)}{k^{n+1}} = \mathcal{O}(k^{-d-1}).
    \label{eq:rahaman_decay}
\end{equation}
\textit{Furthermore, the convergence rate of the residual error $\tilde{h}(k, t)$ at frequency $k$ during training is bounded. As derived in Eq. (9) of \citet{rahaman2019spectralbiasneuralnetworks}:}
\begin{equation}
    \left| \frac{d \tilde{h}(k)}{dt} \right| \propto \mathcal{O}(k \cdot \tilde{f}_\theta(k)) \approx \mathcal{O}(k^{-\Delta}), \quad \text{with } \Delta \ge 1.
    \label{eq:rahaman_convergence}
\end{equation}
\textit{Consequently, for a high-frequency target function $\tau(\mathbf{x})$ (where $\tilde{\tau}(k) \neq 0$ for large $k$), the network output $f_\theta$ collapses to the low-frequency component within any finite training budget.}

\subsection{Proof of Collapse in $\boldsymbol{\epsilon}$-Prediction}
In the standard DDIM setup (Sec.~\ref{sec:appendix_background:ddim}), the network $\boldsymbol{\epsilon}_\theta$ minimizes the MSE against the target noise $\boldsymbol{\epsilon}$. For our input, the target is the normalized high-frequency perturbation: $\boldsymbol{\epsilon} \propto \boldsymbol{\delta}_H$.
The target spectrum is flat (white noise): $|\tilde{\boldsymbol{\epsilon}}(k)| \approx C$.

\textbf{Step 1: Network Output Analysis.}
By Eq.~\eqref{eq:rahaman_decay}, the network $\boldsymbol{\epsilon}_\theta$ cannot fit the flat spectrum of $\boldsymbol{\delta}_H$; it is constrained to decay as $\mathcal{O}(k^{-d-1})$. Furthermore, by Eq.~\eqref{eq:rahaman_convergence}, the learning updates for these high frequencies are negligible. The optimal estimator under these spectral constraints is the conditional expectation:
\begin{equation}
    \boldsymbol{\epsilon}_\theta(\boldsymbol{\mu}_L + \boldsymbol{\delta}_H, y, t) \approx \mathbb{E}[\boldsymbol{\epsilon} \mid \boldsymbol{\mu}_L] \approx \mathbf{0}.
\end{equation}
Substituting this into Eq.~\eqref{eq:ddim_score_from_eps}, the score estimator vanishes: $\nabla_{\mathbf{x}} \log p_t \approx \mathbf{0}$.

\textbf{Step 2: ODE Dynamics Analysis.}
Substituting the vanishing score into the PF-ODE (Eq.~\ref{eq:pf_ode_general}):
\begin{equation}
    \frac{d\mathbf{x}_t}{dt} \approx -\frac{1}{2}\beta(t) \mathbf{x}_t.
\end{equation}
Projecting this ODE onto the high-frequency subspace $\mathcal{H}$ spanned by $\boldsymbol{\delta}_H$:
\begin{equation}
    \frac{d \boldsymbol{\delta}_H(t)}{dt} = -\frac{1}{2}\beta(t) \boldsymbol{\delta}_H(t).
\end{equation}
This describes a \textbf{linear decay process}. The solution is $\boldsymbol{\delta}_H(t) = \boldsymbol{\delta}_H(0) \exp\left(-\frac{1}{2}\int_0^t \beta(s)ds\right)$.
The microscopic noise $\boldsymbol{\delta}_H$ vanishes as $t \to T$. The vector field lacks the restorative component required to amplify variance, leading to spectral collapse.

\subsection{Proof of Robustness in $\mathbf{x}_0$-Prediction}
In the EDM-style parameterization (Sec.~\ref{sec:appendix_background:edm}), the network $F_\theta$ minimizes MSE against the clean data $\mathbf{x}_0$. For our input, the target is the low-frequency signal: $\mathbf{x}_{0} \approx \boldsymbol{\mu}_L$.
The target spectrum decays rapidly: $|\tilde{\mathbf{x}}_{0}(k)| \approx 0$ for large $k$.

\textbf{Step 1: Network Output Analysis.}
The target spectrum aligns with the network's spectral bias (Eq.~\ref{eq:rahaman_decay}). The convergence rate (Eq.~\ref{eq:rahaman_convergence}) is fast for the dominant low-frequencies. The network accurately approximates the projection onto the low-frequency manifold:
\begin{equation}
    F_\theta(\boldsymbol{\mu}_L + \boldsymbol{\delta}_H, y, t) \approx \boldsymbol{\mu}_L.
\end{equation}

\textbf{Step 2: Score Derivation via Residuals.}
The score is derived analytically. Using the definition of the PF-ODE in terms of the denoiser $F_\theta$ (Eq.~\ref{eq:edm_score_from_denoiser}):
\begin{equation}
    \nabla_{\mathbf{x}} \log p_t(\mathbf{x}_t) \approx \frac{F_\theta(\mathbf{x}_t) - \mathbf{x}_t}{\sigma_t^2}.
\end{equation}
Substituting the network output $F_\theta \approx \boldsymbol{\mu}_L$ and input $\mathbf{x}_t \approx \boldsymbol{\mu}_L + \boldsymbol{\delta}_H$:
\begin{equation}
    \nabla_{\mathbf{x}} \log p_t(\mathbf{x}_t) \approx \frac{\boldsymbol{\mu}_L - (\boldsymbol{\mu}_L + \boldsymbol{\delta}_H)}{\sigma_t^2} = -\frac{\boldsymbol{\delta}_H}{\sigma_t^2}.
\end{equation}
Crucially, the high-frequency term $\boldsymbol{\delta}_H$ is recovered via the \textbf{residual connection} $(\mathbf{x}_t - F_\theta)$.

\textbf{Step 3: ODE Dynamics Analysis.}
Substituting this derived score into the general form of the PF-ODE (adapted for EDM parameterization):
\begin{equation}
    \frac{d\mathbf{x}_t}{dt} \propto \text{Drift}(\mathbf{x}_t) - \text{Score}(\mathbf{x}_t) \propto \dots + \frac{\boldsymbol{\delta}_H}{\sigma_t^2}.
\end{equation}
The high-frequency residual creates a positive feedback loop in the ODE vector field (since the score coefficient scales inversely with noise level). This creates a \textbf{repulsive vector field} that exponentially amplifies the microscopic perturbation $\boldsymbol{\delta}_H$, ensuring $\mathbf{x}_T$ converges to full-rank Gaussian noise.

\subsection{Corollaries: Implications for Sampling and Translation}
\label{sec:appendix_theory:corollaries}

Our analysis explains the asymmetry between the failure of inversion on flat inputs and the success of sampling or translation on textured inputs. We verify this via two corollaries.

\textbf{Corollary 1 (Reversibility of Spectral Collapse).}
\textit{Spectral collapse does not prevent perfect reconstruction of the input.}

\textit{Proof.} As derived in Section \ref{sec:appendix_theory} (Step 2), under $\boldsymbol{\epsilon}$-prediction with a flat input, the PF-ODE collapses to the linear decay $\frac{d\mathbf{x}}{dt} = -\frac{1}{2}\beta(t)\mathbf{x}$. The forward map $\Phi$ from $t=0$ to $t=T$ is a scalar contraction:
\begin{equation}
    \mathbf{x}_T = \Phi(\mathbf{x}_0) = \mathbf{x}_0 \exp\left(-\frac{1}{2}\int_0^T \beta(s)ds\right) = C \cdot \mathbf{x}_0.
\end{equation}
While $\mathbf{x}_T$ is rank-deficient (failing the Gaussian prior hypothesis $p_T(\mathbf{x}) \neq \mathcal{N}(\mathbf{0}, \mathbf{I})$), the mapping $\Phi$ remains a linear automorphism (bijective). During reconstruction (backward integration), the network again predicts $\boldsymbol{\epsilon} \approx \mathbf{0}$ given the scaled input $C \cdot \mathbf{x}_0$. The backward ODE becomes a linear expansion, yielding $\hat{\mathbf{x}}_0 = \mathbf{x}_T / C = \mathbf{x}_0$.

\textbf{Corollary 2 (Stability Asymmetry).}
\textit{The inversion is stable if and only if the input spectrum contains high-frequency variance (e.g., standard sampling or high-to-low translation).}

\textit{Proof.}
Let the input be $\mathbf{x}_t = \boldsymbol{\mu}_L + \gamma \boldsymbol{\delta}_H$, where $\gamma$ controls the high-frequency magnitude.
In $\boldsymbol{\epsilon}$-prediction, the network minimizes $\mathbb{E}[\|\boldsymbol{\epsilon} - \boldsymbol{\epsilon}_\theta(\mathbf{x}_t)\|^2]$.
For \textbf{Sampling/High-Freq Translation} ($\gamma \gg 0$): The input $\mathbf{x}_t$ physically contains the noise term $\boldsymbol{\delta}_H$. The target $\boldsymbol{\epsilon}$ is highly correlated with the input component $\boldsymbol{\delta}_H$. The gradient $\nabla_{\mathbf{x}} \boldsymbol{\epsilon}_\theta$ along the direction $\boldsymbol{\delta}_H$ is non-zero, preventing vector field collapse.
Conversely, for \textbf{Flat Inversion} ($\gamma \approx 0$): The correlation vanishes. The network must map $\boldsymbol{\mu}_L \to \boldsymbol{\delta}_H$. This orthogonal mapping is suppressed by the spectral bias (Eq.~\ref{eq:rahaman_decay}), causing the instability.
\section{Training Hyperparameters}
\label{sec:appendix_training_details}

We provide the exact training configurations for all models reported in the main paper.
We utilize the AdamW optimizer for all experiments.
For EDM-based models, we adhere to the preconditioning and noise schedule proposed by Karras et al.~\cite{karras2022elucidatingdesignspacediffusionbased}, calibrating $\sigma_{\text{data}}$ based on the empirical statistics of the latent space for each specific dataset.

\begin{table}[h]
\centering
\caption{\textbf{Detailed Training Hyperparameters.}
We report the optimization settings for each dataset and framework (DDIM vs. EDM).
\textit{Note:} For Super-Resolution (SR) tasks on BBBC021, batch sizes are adjusted (16/32/8) to accommodate GPU memory constraints at different resolutions ($\times8/\times16/\times32$), while keeping optimization parameters consistent.
}
\label{tab:hyperparams}
\resizebox{\textwidth}{!}{%
\begin{tabular}{l|c|cccccc}
\toprule
\textbf{Experiment / Dataset} & \textbf{Framework} & \textbf{Effective Batch Size} & \textbf{Learning Rate} & \textbf{LR Schedule} & \textbf{EMA Decay} & \textbf{Steps} & \textbf{EDM Parameters} \\
\midrule
\multicolumn{8}{l}{\textit{\textbf{Pixel-Space Experiments}}} \\
\midrule
\multirow{2}{*}{BBBC021 (Pixel)} 
 & DDIM & 64 & $5\times10^{-5}$ & Constant & 0.995 & 80k & N/A \\
 & EDM & 64 & $1\times10^{-4}$ & Cosine & 0.999 & 100k & $\begin{aligned} \sigma_{\text{data}} &= 0.5 \\ P_{\text{mean}} &= -1.2 \end{aligned}$ \\
\midrule
\multicolumn{8}{l}{\textit{\textbf{Latent-Space Experiments}}} \\
\midrule
\multirow{2}{*}{Edges2Shoes} 
 & DDIM & 128 & $5\times10^{-5}$ & Cosine & 0.995 & 80k & N/A \\
 & EDM & 128 & $5\times10^{-5}$ & Constant & 0.995 & 80k & $\begin{aligned} \sigma_{\text{data}} &= 0.7 \\ P_{\text{mean}} &= -0.4 \end{aligned}$ \\
\midrule
\multirow{2}{*}{BBBC021 SR ($\times8, \times16, \times32$)} 
 & DDIM & 128 & $5\times10^{-5}$ & Cosine & 0.995 & 80k & N/A \\
 & EDM & 128 & $5\times10^{-5}$ & Cosine & 0.995 & 80k & $\begin{aligned} \sigma_{\text{data}} &= 1.14 \\ P_{\text{mean}} &= -1.89 \end{aligned}$ \\
\midrule
\multicolumn{8}{l}{\textit{\textbf{Ablations (Architecture)}}} \\
\midrule
\multirow{2}{*}{BBBC021 (DiT-L/2)} 
 & DDIM & 128 & $5\times10^{-5}$ & Cosine & 0.995 & 120k & N/A \\
 & EDM & 128 & $5\times10^{-5}$ & Cosine & 0.995 & 120k & $\begin{aligned} \sigma_{\text{data}} &= 0.5 \\ P_{\text{mean}} &= -1.2 \end{aligned}$ \\
\bottomrule
\end{tabular}%
}
\footnotesize{\textbf{Standard EDM Constants:} Across all EDM experiments, we set $\sigma_{\min}=0.002$, $\sigma_{\max}=80.0$, and $P_{\text{std}}=1.2$.}
\vspace{2mm}
\end{table}

\textbf{Hardware \& Optimization:} All models were trained on 4 NVIDIA A100 GPUs with a batch size of 16 and gradient accumulation of 2 (and of 1 for \textit{pixel} space).

\section{Additional Results}

\subsection{Additional Correlation Plots between Scores}
\label{sec:additional_correlation_plots}

\begin{figure}[H]
    \centering
    \includegraphics[width=0.45\columnwidth]{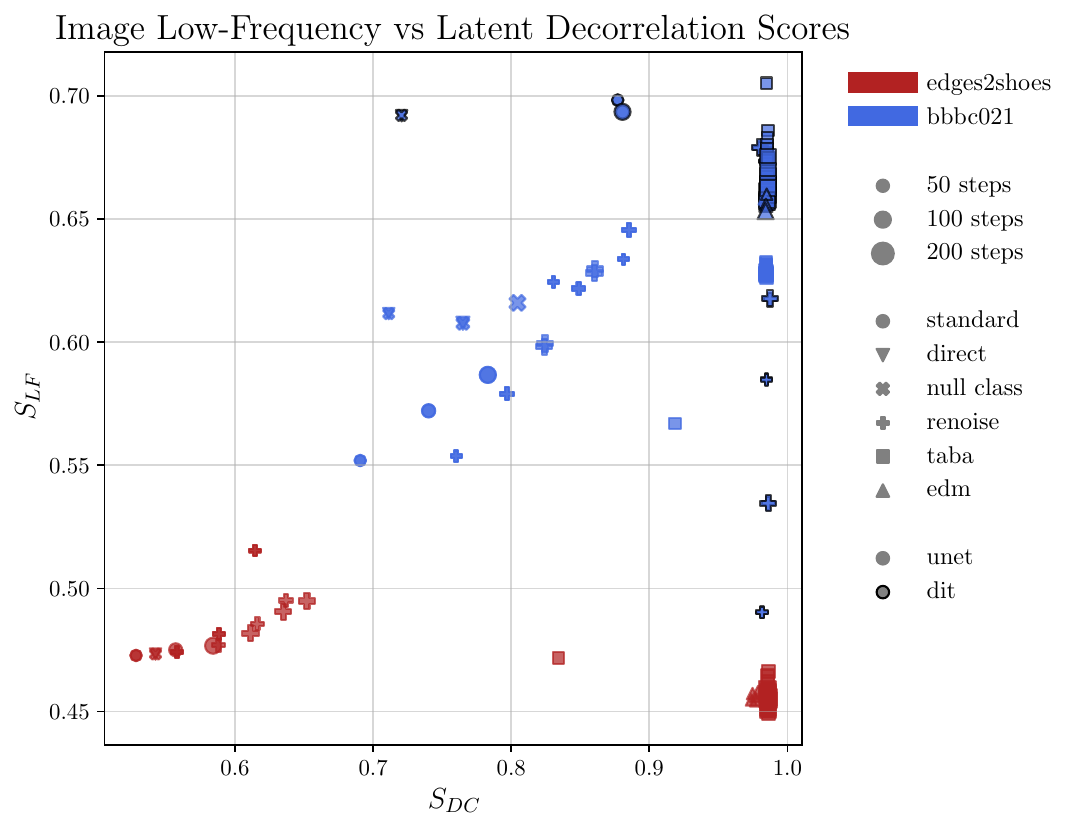}
    \caption{Variation of the Low-Frequency Score $S_{\mathrm{LF}}$ against the Latent Decorrelation Score $S_{\mathrm{DC}}$ across different methods and datasets. A moderate positive correlation is observed, indicating that as the patch correlation in the latent increases (higher spectral collapse), the structural fidelity of the generated images to the input condition improves slightly.}
    \label{app:fig:slf_vs_sdc}
\end{figure}

\subsection{Velocity Heatmaps: Spectral Dynamics of Inversion}
\label{sec:appendix_velocity_heatmaps}

\begin{figure}[ht]
    \centering
    \includegraphics[width=\linewidth]{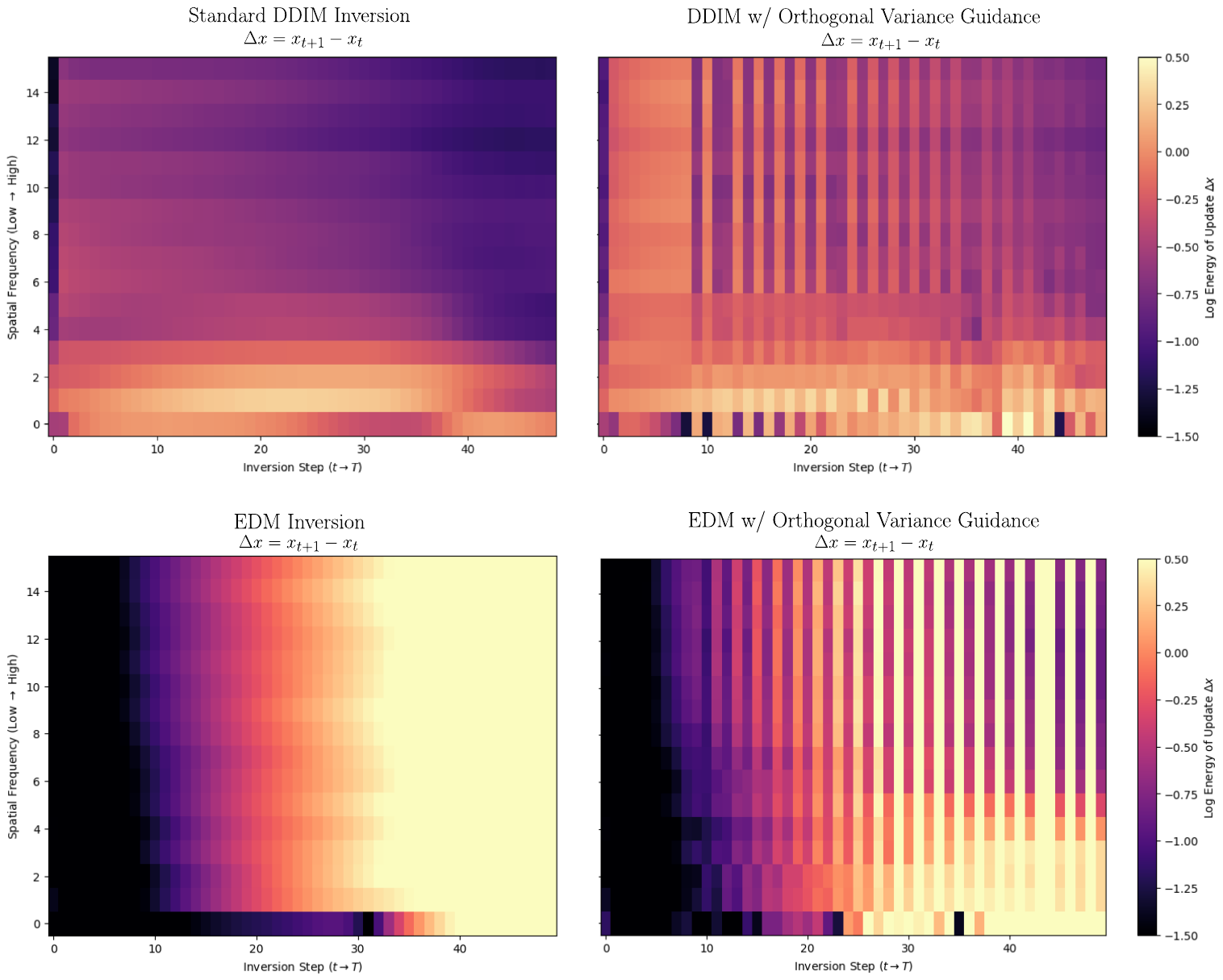}
    \caption{\textbf{Spectral Velocity Analysis}
        We visualize the logarithmic energy spectrum of the update vector $\Delta \mathbf{z}_t$ across 50 inversion steps for a single sample from BBBC021 ($\times16$) using a U-Net in latent space.
        \textbf{Top Left} Standard DDIM exhibits negative log-energy regions (dark blue) for high frequencies as $t \to T$. This signifies \textit{vanishing} updates where the solver stops refining texture details due to SNR decay, causing spectral collapse.
        \textbf{Bottom Left} EDM inversion maintains positive update energy throughout, preserving spectral continuity.
        \textbf{Right Column} OVG prevents vanishing updates by actively injecting orthogonal broadband energy (vertical striations), ensuring the latent trajectory remains spectrally active.}
    \label{fig:velocity_heatmaps}
\end{figure}

To dissect the dynamics of spectral collapse, we analyze the spectral energy of the latent updates throughout the inversion trajectory.
We define the "inversion velocity" as the magnitude of the update vector between consecutive steps: $\Delta \mathbf{z}_t = \mathbf{z}_{t+1} - \mathbf{z}_t$.
Figure~\ref{fig:velocity_heatmaps} visualizes the Fourier transform magnitude of $\Delta \mathbf{z}_t$ (averaged over spatial dimensions) on a logarithmic scale.
In these heatmaps, "negative values" (dark blue regions) correspond to $\log |\Delta \mathbf{z}_t| \ll 0$, indicating that the update magnitude is asymptotically approaching zero.
This analysis reveals a critical divergence in solver behavior:
\paragraph{Standard DDIM (Vanishing High-Frequency Updates)} The standard DDIM heatmap (top left) is characterized by a distinct "dark zone" of negative log-values in the high-frequency band as $t \to T$. We interpret this phenomenon as progressively \textit{vanishing} updates: as the Signal-to-Noise Ratio (SNR) degrades towards the end of the inversion, the $\boldsymbol{\epsilon}$-prediction model fails to resolve high-frequency structural details from the noise. Consequently, the ODE solver effectively ceases to modify these frequencies (update magnitude $\approx 0$), "freezing" the texture in an unresolved state.
\paragraph{EDM Inversion (Spectral Continuity)} In contrast, EDM inversion (bottom left) avoids these signal voids. The heatmap displays consistent, positive log-energy across all frequencies throughout the entire trajectory. By parameterizing the network to predict $\mathbf{x}_0$ directly and employing spectrally balanced preconditioning, EDM maintains non-vanishing gradients even in low-SNR regimes. This ensures that high-frequency texture is continuously refined rather than frozen.
\paragraph{Orthogonal Variance Guidance} the application of Orthogonal Variance Guidance (right column) introduces visible vertical striations. These represent active injections of orthogonal energy that prevent the log-energy from dipping into the negative (vanishing) range. For DDIM, OVG acts as a "spectral guide" forcing the solver to maintain a non-zero velocity in the orthogonal subspace and thereby preventing the high-frequency stagnation observed in the baseline.

\subsection{Additional Qualitative Results}
\label{sec:appendix_qualitative}

To demonstrate the versatility and robustness of our approach, we provide extended qualitative results on BBBC021 \& Edges2shoes datasets across a diverse set of experimental conditions.
We evaluate performance across varying degrees of spectral sparsity (Super-Resolution factors $\times8, \times16, \times32$), different diffusion training objectives ($\boldsymbol{\epsilon}$-prediction vs. $\mathbf{x}_0$-prediction), and distinct backbone architectures (U-Net vs. DiT).
Consistent with our main analysis, we observe that deterministic baselines suffer from increasingly severe spectral collapse as the task difficulty increases (e.g., at $\times32$ resolution). In contrast, our Orthogonal Variance Guidance (OVG) restores plausible high frequencies \& textures across all configurations.

\begin{figure}[ht]
    \centering
    \includegraphics[
        width=0.8\columnwidth]{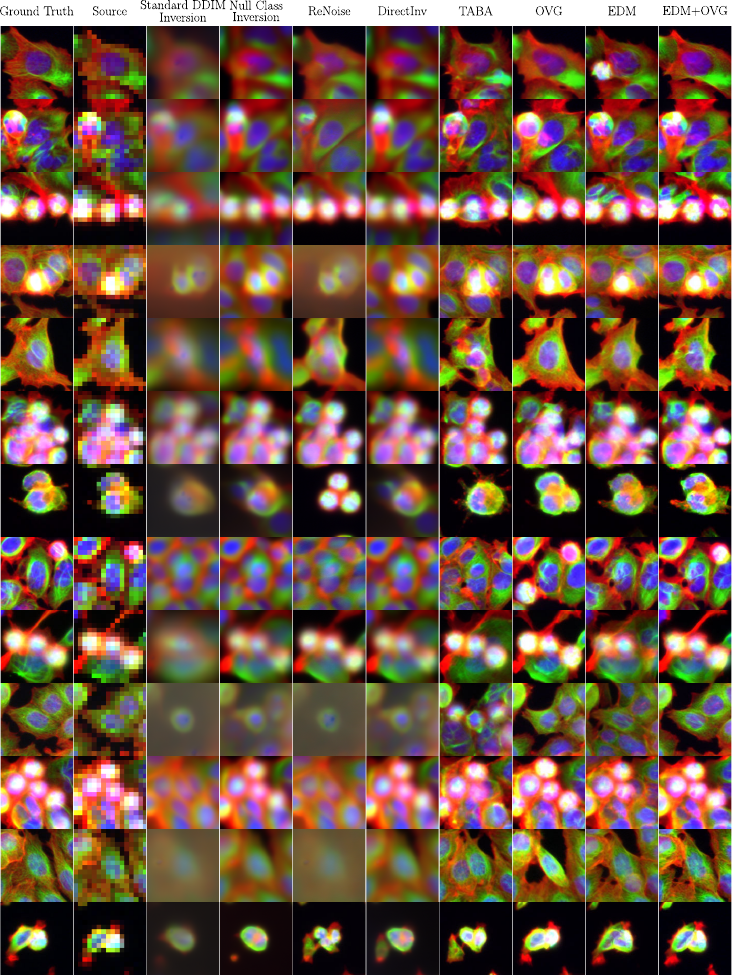}
    \caption{\textbf{Qualitative Comparison on BBBC021 x16 UNet Latent $\boldsymbol{\epsilon}$ pred}
        We super-resolve microscopy images from downsampled inputs (source).
        \textbf{Deterministic methods} (Standard DDIM, DirectInv, Null-Class) exhibit severe \textit{spectral collapse}, yielding over-smoothed, blurry predictions that fail to reconstruct cellular details.
        \textbf{Stochastic methods} (ReNoise, TABA) inject high-frequency content but often suffer from incoherent hallucinations or noise artifacts.
        \textbf{OVG \& EDM+OVG} achieve superior fidelity, reconstructing sharp biological structures (nuclei, membranes) that are both texturally realistic and spatially aligned with the source.}
    \label{fig:bbbc021_unet_latent_epsilon_qualitative}
\end{figure}

\begin{figure}[ht]
    \centering
    \includegraphics[width=0.9\columnwidth]{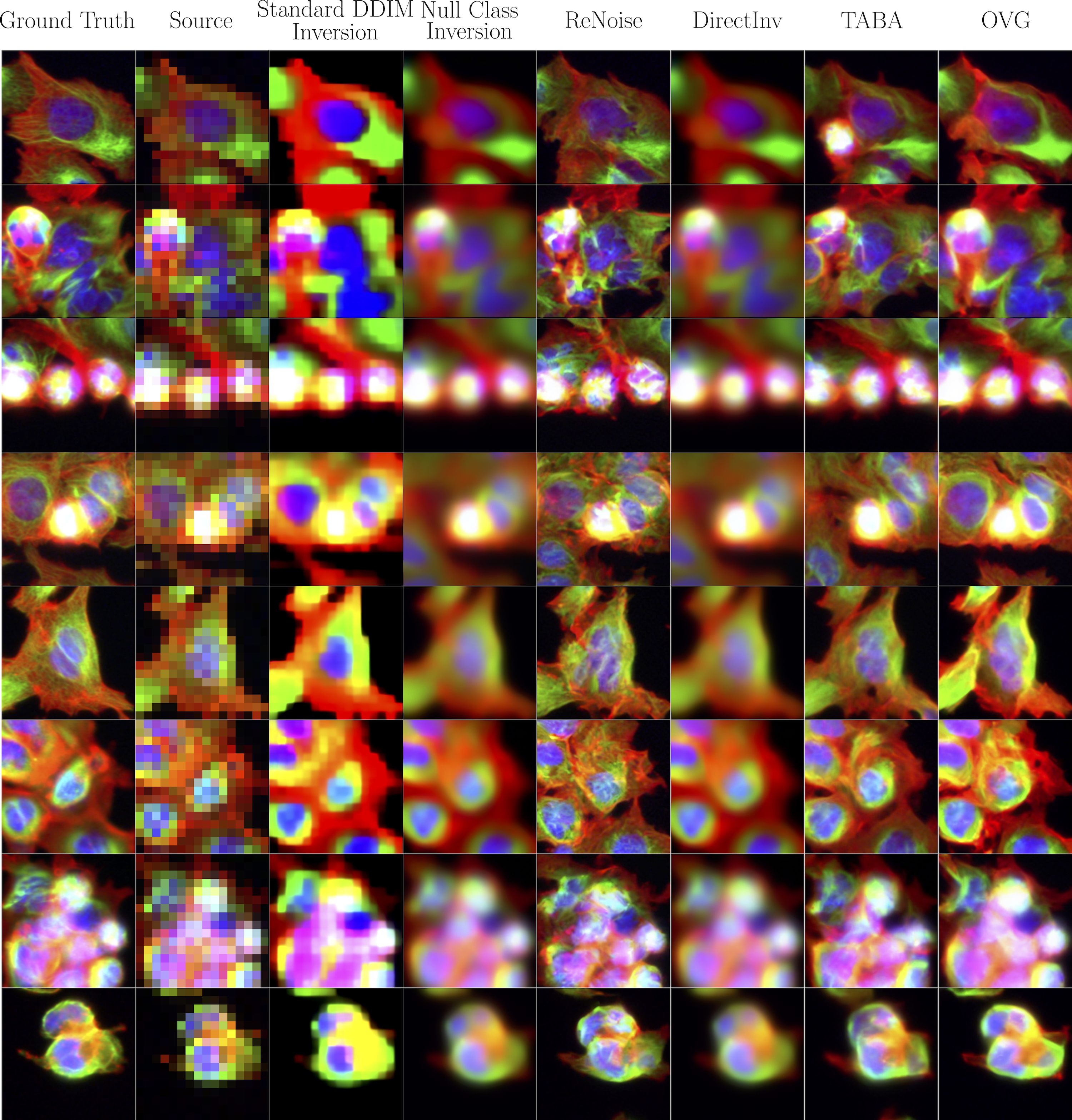}
    \caption{\textbf{Qualitative Comparison on BBBC021 x16 DiT Latent $\boldsymbol{\epsilon}$ pred}
        Using a Transformer-based diffusion model (DiT), we observe the same spectral collapse phenomenon: \textbf{deterministic methods} (Standard DDIM, Null-Class, DirectInv) yield blurry, mean-like predictions. While \textbf{stochastic baselines} (ReNoise, TABA) recover some texture, they struggle with consistency. \textbf{OVG} successfully synthesizes plausible high-frequency biological details while maintaining the structural integrity of the low-resolution input}
    \label{fig:bbbc021_dit_latent_epsilon_qualitative}
\end{figure}

\begin{figure}[ht]
    \centering
    \includegraphics[width=0.95\columnwidth]{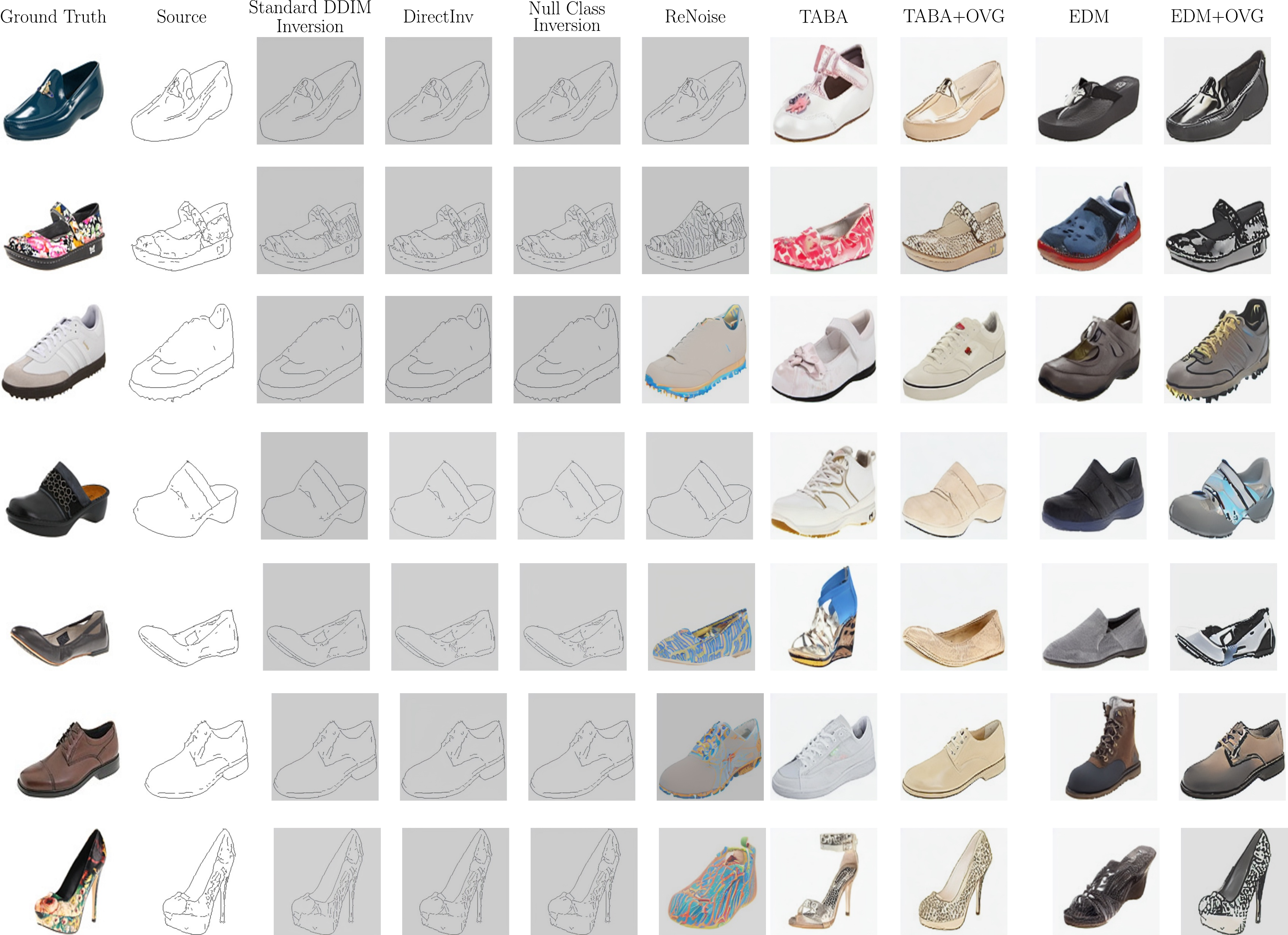}
    \caption{\textbf{Qualitative Comparison on Edges2Shoes}
        We translate edge maps (source) to shoes. \textbf{Deterministic methods} (DDIM, DirectInv, Null-Class) suffer from severe \textit{spectral collapse}, producing gray, texture-less outputs locked by poisoned inversion.
        \textbf{Stochastic methods} (TABA) restore texture but may introduce artifacts and drifts from source structure.
        \textbf{OVG \& EDM+OVG} achieve the best perceptual quality, generating realistic textures while maintaining precise alignment with the input edges.}
    \label{fig:edges2shoes_unet_latent_qualitative}
\end{figure}

\subsection{Additional Quantitative Results}
In this section, we provide the complete quantitative results for all methods and datasets evaluated in our study. We provide the performance metrics across BBBC021 x8, x16 and x32 and Edges2Shoes datasets.

\begin{table}[H]
    \centering
    \caption{BBBC021 in x8 resolution with 50 inversion steps. Standard deviation over 3 seeds reported.}
    \setlength{\tabcolsep}{3pt} 
    \resizebox{\textwidth}{!}{
        \begin{tabular}{l | c c c c c c c}
    \toprule
    \textbf{Method} & \textbf{PSNR} & \textbf{MS-SSIM} & \textbf{LPIPS} & \textbf{FID} & \textbf{$S_{HF}$} & \textbf{$S_{LF}$} & \textbf{$S_{DC}$} \\
    \midrule
    \textit{Deterministic} &  &  &  &  &  &  &  \\
    Standard & 18.08 $\pm$ 0.07 & 0.733 $\pm$ 0.002 & 0.269 $\pm$ 0.002 & 32.41 $\pm$ 0.45 & 0.93 $\pm$ 0.00 & 0.75 $\pm$ 0.00 & 0.97 $\pm$ 0.00 \\
    Null Class & \textbf{19.59 $\pm$ 0.07} & \textbf{0.782 $\pm$ 0.002} & 0.245 $\pm$ 0.001 & 64.62 $\pm$ 0.49 & 0.90 $\pm$ 0.00 & \textbf{0.78 $\pm$ 0.00} & 0.95 $\pm$ 0.00 \\
    Direct & \textbf{19.59 $\pm$ 0.07} & \textbf{0.782 $\pm$ 0.002} & 0.245 $\pm$ 0.001 & 64.63 $\pm$ 0.47 & 0.90 $\pm$ 0.00 & \textbf{0.78 $\pm$ 0.00} & 0.95 $\pm$ 0.00 \\
    \midrule
    \textit{Stochastic} &  &  &  &  &  &  &  \\
    Renoise & 17.08 $\pm$ 0.04 & 0.660 $\pm$ 0.003 & 0.293 $\pm$ 0.002 & 27.01 $\pm$ 0.98 & \textbf{0.97 $\pm$ 0.00} & 0.71 $\pm$ 0.00 & \textbf{0.99 $\pm$ 0.00} \\
    TABA & 18.05 $\pm$ 0.04 & 0.724 $\pm$ 0.002 & 0.256 $\pm$ 0.001 & 17.00 $\pm$ 0.53 & 0.94 $\pm$ 0.00 & 0.74 $\pm$ 0.00 & 0.98 $\pm$ 0.00 \\
    \midrule
    \textit{Ours} &  &  &  &  &  &  &  \\
    EDM & 18.52 $\pm$ 0.08 & 0.729 $\pm$ 0.002 & 0.238 $\pm$ 0.002 & \textbf{15.31 $\pm$ 0.39} & 0.96 $\pm$ 0.00 & 0.75 $\pm$ 0.00 & 0.98 $\pm$ 0.00 \\
    OVG & 18.77 $\pm$ 0.02 & 0.772 $\pm$ 0.001 & \textbf{0.192 $\pm$ 0.000} & \textbf{15.61 $\pm$ 0.03} & 0.96 $\pm$ 0.00 & 0.77 $\pm$ 0.00 & 0.98 $\pm$ 0.00 \\
    OVG + TABA & 18.92 $\pm$ 0.02 & 0.770 $\pm$ 0.001 & 0.194 $\pm$ 0.001 & \textbf{15.42 $\pm$ 0.08} & 0.96 $\pm$ 0.00 & \textbf{0.78 $\pm$ 0.00} & 0.98 $\pm$ 0.00 \\
    OVG + EDM & 19.06 $\pm$ 0.04 & 0.743 $\pm$ 0.002 & 0.220 $\pm$ 0.001 & 18.32 $\pm$ 0.51 & 0.95 $\pm$ 0.00 & \textbf{0.78 $\pm$ 0.00} & 0.97 $\pm$ 0.00 \\
    \bottomrule
\end{tabular}}
\end{table}

\begin{table}[H]
    \centering
    \caption{BBBC021 in x16 resolution with 50 inversion steps. Standard deviation over 3 seeds reported.}
    \setlength{\tabcolsep}{3pt} 
    \resizebox{\textwidth}{!}{
        \begin{tabular}{l | c c c c c c c}
    \toprule
    \textbf{Method} & \textbf{PSNR} & \textbf{MS-SSIM} & \textbf{LPIPS} & \textbf{FID} & \textbf{$S_{HF}$} & \textbf{$S_{LF}$} & \textbf{$S_{DC}$} \\
    \midrule
    \textit{Deterministic} &  &  &  &  &  &  &  \\
    Standard & 15.11 $\pm$ 0.08 & 0.553 $\pm$ 0.003 & 0.521 $\pm$ 0.002 & 217.09 $\pm$ 2.25 & 0.83 $\pm$ 0.00 & 0.55 $\pm$ 0.00 & 0.69 $\pm$ 0.00 \\
    Null Class & 16.22 $\pm$ 0.05 & 0.616 $\pm$ 0.002 & 0.434 $\pm$ 0.000 & 186.78 $\pm$ 1.97 & 0.85 $\pm$ 0.00 & 0.61 $\pm$ 0.00 & 0.71 $\pm$ 0.00 \\
    Direct & 16.22 $\pm$ 0.05 & 0.616 $\pm$ 0.002 & 0.434 $\pm$ 0.000 & 186.76 $\pm$ 1.95 & 0.85 $\pm$ 0.00 & 0.61 $\pm$ 0.00 & 0.71 $\pm$ 0.00 \\
    \midrule
    \textit{Stochastic} &  &  &  &  &  &  &  \\
    Renoise & 16.20 $\pm$ 0.10 & 0.606 $\pm$ 0.005 & 0.457 $\pm$ 0.005 & 145.57 $\pm$ 2.74 & 0.92 $\pm$ 0.00 & 0.62 $\pm$ 0.00 & 0.83 $\pm$ 0.00 \\
    TABA & 15.90 $\pm$ 0.01 & 0.600 $\pm$ 0.004 & 0.329 $\pm$ 0.002 & \textbf{12.06 $\pm$ 0.02} & 0.96 $\pm$ 0.00 & 0.63 $\pm$ 0.00 & \textbf{0.98 $\pm$ 0.00} \\
    \midrule
    \textit{Ours} &  &  &  &  &  &  &  \\
    EDM & \textbf{16.92 $\pm$ 0.05} & 0.634 $\pm$ 0.002 & 0.294 $\pm$ 0.000 & 14.21 $\pm$ 0.35 & 0.96 $\pm$ 0.00 & 0.66 $\pm$ 0.00 & 0.98 $\pm$ 0.00 \\
    OVG & 16.66 $\pm$ 0.01 & \textbf{0.665 $\pm$ 0.003} & 0.281 $\pm$ 0.001 & 18.62 $\pm$ 0.00 & 0.95 $\pm$ 0.00 & \textbf{0.67 $\pm$ 0.00} & 0.97 $\pm$ 0.00 \\
    OVG + TABA & 16.54 $\pm$ 0.03 & 0.656 $\pm$ 0.002 & \textbf{0.279 $\pm$ 0.001} & 13.14 $\pm$ 0.08 & 0.96 $\pm$ 0.00 & \textbf{0.67 $\pm$ 0.00} & 0.98 $\pm$ 0.00 \\
    OVG + EDM & \textbf{16.90 $\pm$ 0.04} & 0.621 $\pm$ 0.002 & 0.299 $\pm$ 0.001 & 17.37 $\pm$ 0.39 & \textbf{0.96 $\pm$ 0.00} & 0.65 $\pm$ 0.00 & 0.98 $\pm$ 0.00 \\
    \bottomrule
\end{tabular}}
\end{table}
\begin{table}[H]
    \centering
    \caption{BBBC021 in x32 resolution with 50 inversion steps. Standard deviation over 3 seeds reported.}
    \setlength{\tabcolsep}{3pt} 
    \resizebox{\textwidth}{!}{
        \begin{tabular}{l | c c c c c c c}
    \toprule
    \textbf{Method} & \textbf{PSNR} & \textbf{MS-SSIM} & \textbf{LPIPS} & \textbf{FID} & \textbf{$S_{HF}$} & \textbf{$S_{LF}$} & \textbf{$S_{DC}$} \\
    \midrule
    \textit{Deterministic} &  &  &  &  &  &  &  \\
    Standard & \textbf{14.58 $\pm$ 0.05} & \textbf{0.529 $\pm$ 0.002} & 0.459 $\pm$ 0.002 & 99.14 $\pm$ 1.85 & 0.90 $\pm$ 0.00 & \textbf{0.58 $\pm$ 0.00} & 0.88 $\pm$ 0.00 \\
    Null Class & 14.44 $\pm$ 0.04 & 0.486 $\pm$ 0.003 & 0.575 $\pm$ 0.004 & 258.42 $\pm$ 3.52 & 0.76 $\pm$ 0.00 & 0.53 $\pm$ 0.00 & 0.65 $\pm$ 0.00 \\
    Direct & 14.44 $\pm$ 0.04 & 0.486 $\pm$ 0.003 & 0.575 $\pm$ 0.004 & 258.42 $\pm$ 3.51 & 0.76 $\pm$ 0.00 & 0.53 $\pm$ 0.00 & 0.65 $\pm$ 0.00 \\
    \midrule
    \textit{Stochastic} &  &  &  &  &  &  &  \\
    Renoise & 13.59 $\pm$ 0.04 & 0.451 $\pm$ 0.003 & 0.451 $\pm$ 0.001 & 22.22 $\pm$ 0.52 & 0.97 $\pm$ 0.00 & 0.53 $\pm$ 0.00 & 0.97 $\pm$ 0.00 \\
    TABA & 14.20 $\pm$ 0.02 & 0.495 $\pm$ 0.003 & \textbf{0.422 $\pm$ 0.001} & \textbf{14.41 $\pm$ 0.25} & 0.95 $\pm$ 0.00 & 0.55 $\pm$ 0.00 & \textbf{0.99 $\pm$ 0.00} \\
    \midrule
    \textit{Ours} &  &  &  &  &  &  &  \\
    EDM & \textbf{14.58 $\pm$ 0.05} & \textbf{0.529 $\pm$ 0.002} & 0.459 $\pm$ 0.002 & 99.13 $\pm$ 1.85 & 0.90 $\pm$ 0.00 & \textbf{0.58 $\pm$ 0.00} & 0.88 $\pm$ 0.00 \\
    OVG & 13.89 $\pm$ 0.04 & 0.484 $\pm$ 0.003 & \textbf{0.423 $\pm$ 0.001} & 15.31 $\pm$ 0.34 & \textbf{0.97 $\pm$ 0.00} & 0.53 $\pm$ 0.00 & 0.97 $\pm$ 0.00 \\
    OVG + TABA & 13.91 $\pm$ 0.04 & 0.486 $\pm$ 0.003 & \textbf{0.424 $\pm$ 0.001} & 15.65 $\pm$ 0.25 & 0.96 $\pm$ 0.00 & 0.53 $\pm$ 0.00 & 0.98 $\pm$ 0.00 \\
    OVG + EDM & 13.89 $\pm$ 0.04 & 0.484 $\pm$ 0.003 & \textbf{0.423 $\pm$ 0.001} & 15.34 $\pm$ 0.35 & \textbf{0.97 $\pm$ 0.00} & 0.53 $\pm$ 0.00 & 0.97 $\pm$ 0.00 \\
    \bottomrule
\end{tabular}}
\end{table}
\begin{table}[H]
    \centering
    \caption{Edges2Shoes dataset with 50 inversion steps. Standard deviation over 3 seeds reported.}
    \setlength{\tabcolsep}{3pt} 
    \resizebox{\textwidth}{!}{
        \begin{tabular}{l | c c c c c c c}
    \toprule
    \textbf{Method} & \textbf{PSNR} & \textbf{MS-SSIM} & \textbf{LPIPS} & \textbf{FID} & \textbf{$S_{HF}$} & \textbf{$S_{LF}$} & \textbf{$S_{DC}$} \\
    \midrule
    \textit{Deterministic} &  &  &  &  &  &  &  \\
    Standard & 7.42 $\pm$ 0.00 & 0.404 $\pm$ 0.000 & 0.467 $\pm$ 0.000 & 385.01 $\pm$ 0.05 & 0.69 $\pm$ 0.00 & 0.47 $\pm$ 0.00 & 0.53 $\pm$ 0.00 \\
    Null Class & 7.41 $\pm$ 0.00 & 0.403 $\pm$ 0.000 & 0.466 $\pm$ 0.000 & 392.73 $\pm$ 0.06 & 0.69 $\pm$ 0.00 & 0.47 $\pm$ 0.00 & 0.54 $\pm$ 0.00 \\
    Direct & 7.41 $\pm$ 0.00 & 0.403 $\pm$ 0.000 & 0.466 $\pm$ 0.000 & 392.82 $\pm$ 0.01 & 0.69 $\pm$ 0.00 & 0.47 $\pm$ 0.00 & 0.54 $\pm$ 0.00 \\
    \midrule
    \textit{Stochastic} &  &  &  &  &  &  &  \\
    Renoise & 7.95 $\pm$ 0.00 & 0.423 $\pm$ 0.000 & 0.462 $\pm$ 0.000 & 305.31 $\pm$ 0.03 & 0.74 $\pm$ 0.00 & 0.52 $\pm$ 0.00 & 0.61 $\pm$ 0.00 \\
    TABA & 8.79 $\pm$ 0.06 & 0.398 $\pm$ 0.003 & 0.463 $\pm$ 0.002 & 93.32 $\pm$ 5.31 & 0.92 $\pm$ 0.01 & 0.46 $\pm$ 0.00 & \textbf{0.99 $\pm$ 0.00} \\
    \midrule
    \textit{Ours} &  &  &  &  &  &  &  \\
    EDM & \textbf{9.59 $\pm$ 0.01} & 0.420 $\pm$ 0.002 & 0.440 $\pm$ 0.001 & \textbf{53.10 $\pm$ 0.36} & \textbf{0.93 $\pm$ 0.00} & 0.46 $\pm$ 0.00 & 0.97 $\pm$ 0.00 \\
    OVG & 8.58 $\pm$ 0.03 & 0.427 $\pm$ 0.002 & 0.433 $\pm$ 0.002 & 145.86 $\pm$ 5.30 & 0.87 $\pm$ 0.01 & 0.54 $\pm$ 0.00 & 0.98 $\pm$ 0.00 \\
    OVG + TABA & 8.49 $\pm$ 0.04 & \textbf{0.436 $\pm$ 0.008} & 0.424 $\pm$ 0.007 & 162.59 $\pm$ 7.75 & 0.85 $\pm$ 0.01 & 0.56 $\pm$ 0.01 & 0.98 $\pm$ 0.00 \\
    OVG + EDM & 9.38 $\pm$ 0.04 & \textbf{0.443 $\pm$ 0.001} & \textbf{0.381 $\pm$ 0.001} & 90.16 $\pm$ 1.56 & 0.90 $\pm$ 0.00 & \textbf{0.60 $\pm$ 0.00} & 0.90 $\pm$ 0.00 \\
    \bottomrule
\end{tabular}}
\end{table}

\end{document}